\documentclass[11pt,letterpaper]{article}
\usepackage{hyperref}
\usepackage[margin=1in]{geometry}
\usepackage{amssymb,amsthm,amsmath,amssymb,wrapfig,dsfont,authblk,mathrsfs}
\usepackage{times}
\usepackage{enumitem}
\usepackage{amsfonts}
\usepackage{comment}
\usepackage{framed}
\usepackage{soul}
\usepackage{thmtools,thm-restate}
\usepackage{csquotes}
\usepackage[nameinlink, noabbrev, capitalize]{cleveref}
\usepackage{microtype}
\usepackage{placeins}
\usepackage{natbib}
\usepackage{dcmacro}
\usepackage{booktabs} %
\usepackage[ruled]{algorithm2e} %

\SetAlFnt{\small}
\SetAlCapFnt{\small}
\SetAlCapNameFnt{\small}
\SetAlCapHSkip{0pt}
\IncMargin{-\parindent}

\usepackage{graphicx}
\usepackage{multirow,hhline,tabularx,xcolor,wrapfig,tikz}
\usepackage{thm-restate}

\begin{document}
\title{Distributional Adversarial Loss\footnote{Authors are ordered alphabetically.}}
\author{Saba Ahmadi}
\author{Siddharth Bhandari}
\author{Avrim Blum}
\author{Chen Dan}
\author{Prabhav Jain} 
\affil{Toyota Technological Institute at Chicago\\
{\small\texttt{\{saba,siddharth,avrim,chendan,prabhavjain\}@ttic.edu}}}

\maketitle

\begin{abstract}

We initiate the study of a new notion of adversarial loss which we call \emph{distributional adversarial loss}. In this notion, we assume for each original example, the allowed adversarial perturbation set is a family of \emph{distributions}, and the adversarial loss over each example is the maximum loss over all the associated distributions. The goal is to minimize the overall adversarial loss. We show sample complexity bounds in the PAC-learning setting for our notion of adversarial loss. Our notion of adversarial loss contrasts the prior work on robust learning that considers a set of points, not distributions, as the perturbation set of each clean example. As an application of our approach, we show how to unify the two lines of work on randomized smoothing and robust learning in the PAC-learning setting and derive sample complexity bounds for randomized smoothing methods.

Furthermore, we investigate the role of randomness in achieving robustness against adversarial attacks. We show a general derandomization technique that preserves the extent of a randomized classifier's robustness against adversarial attacks and show its effectiveness empirically.
\end{abstract}

\section{Introduction and Related Work}
Recent research extensively explores the development of robust predictors against adversarial perturbations, revealing the susceptibility of deep neural networks to imperceptible adversarial noise~\cite{biggio2013evasion, szegedy2013intriguing, goodfellow2014explaining}. Adversarial perturbations involve introducing limited noise $\delta$ to an image ${x}$ (or more generally, $x'\in \calA(x)$ for a perturbation set $\calA(x)$), resulting in visually indistinguishable yet misclassified instances. This phenomenon poses significant threats to real-world applications such as self-driving cars~\cite{cao2021invisible} and healthcare~\cite{finlayson2019adversarial}. To bolster classifier resilience against these perturbations, various empirical defenses have been proposed, %
however, these methods often overfit and still exhibit vulnerability to meticulously crafted adversaries on test points~\citep{carlini2017adversarial, athalye2018robustness}.

Subsequently, techniques for \emph{certifiable robustness} have been introduced. These methods ensure that for any given input ${x}$, whether it is clean or perturbed, a radius $\rho$ can be determined such that all inputs ${x'}$ within the distance $\rho$ from the original input ${x}$ are guaranteed to receive the same label as ${x}$%
. \emph{Randomized smoothing} methods~\citep{cohen2019certified,salman2019provably,zhai2020macer,mohapatra2020higher} have been proposed as %
certifiable robustness techniques that scale to large-scale datasets such as ImageNet. In randomized smoothing, given a possibly perturbed input, the final classification is provided by taking the majority vote over the Gaussian-smoothed perturbations of the input, or other suitable smoothing perturbations. The principle here is that the added noise helps to drown out the adversarial perturbation present in the input and counteract an adversary's power.%

In another line of work, the sample complexity of robust learning in the PAC-learning setting has been studied, e.g. ~\citet{cullina2018pac,attias2022improved,montasser2019vc}, where %
the goal is to learn a predictor having small robust loss that is defined as $\Ex_{(x,y)\sim \calD} \insquare{ \max_{z\in \calA(x)} \ind\insquare{h(z)\neq y} }$. That is, for each non-adversarial example sampled from an underlying distribution $\calD$, the adversarial loss on $(x,y)$ is defined as the maximum loss over all its perturbations in $\calA(x)$.

Motivated by these two lines of work, %
we propose a new notion of adversarial loss which we call \emph{distributional adversarial loss}. Here, for each example ${x}$, the perturbation set $\calU(x)$ is a family of \emph{distributions} instead of a set of points. A perturbation ${x'}$ of ${x}$ can be sampled from any of the distributions in $\calU(x)$. $\calU(x)$ corresponds roughly to the set of distributions that arise when the adversary chooses a perturbed version of $x$, say $x'$, and $x'$ is smoothed by adding noise %
to it. Our notion contrasts with prior
work that considers a set of points, not distributions, as the perturbation set of each clean example.
Given that, the distributional adversarial loss is defined as $\Ex_{(x,y)\sim \calD} \insquare{ \max_{u\in \calU(x)}\insquare{\Ex_{z\sim u} \ind\insquare{h(z)\neq y}} }$. That is, for each non-adversarial example sampled from an underlying distribution $\calD$, the adversarial loss on $(x,y)$ is defined as the maximum expected loss over all the distributions in its perturbation set $\calU(x)$. The %
population loss is the expected adversarial loss over all the samples coming from the underlying distribution $\calD$. We show PAC-learning results for this notion of adversarial loss.

We expand on two important and desirable properties of optimizing our notion of distributional adversarial loss. First, using our framework, we can derive sample complexity bounds for randomized smoothing methods (\cref{sec:connect-RS-PAC}). Another benefit of our approach is that in some settings, it happens that for each input $x$, any adversarial choice $x'\in \calA(x)$ is covered by some distribution $u\in \calU(x)$\footnote{By covering, we mean that $u$ %
is close in some notion of distance, e.g. total variation distance, to a distribution supported on a ball around $x'$. The radius of the ball depends on the adversary's power. More details in~\Cref{sec:connect-RS-PAC}.}. Therefore, by having a low distributional adversarial loss on $u$, the classifier would also have a good performance on $x'$. In some settings, including randomized smoothing, we can consider a distribution set $\calU(x)$ such that $\calU(x)$ is much smaller than $\calA(x)$.
As a result, instead of considering an \emph{unbounded} number of adversary's actions, it suffices to consider a \emph{bounded} number of distributions that cover the adversary's actions. Another way of viewing this is that now \emph{the same adversary with the same power} has a smaller set of \emph{effective} actions that we need to worry about, a bounded number of distributions that cover an unbounded number of perturbation points. %

\paragraph{Derandomization:} Another natural point of inquiry about robust classifiers (certifiable or not) built using the above approaches is whether they can be made deterministic, i.e., no randomness is required during inference time. This question has been answered in the affirmative in some cases by~\cite{levine2021improved}. There are a few advantages to having a deterministic classifier. In pivotal decisions, we aim for the classifier's output to be deterministic, ensuring consistent labeling for a given input (likewise for the certification radius). Additionally, derandomization has the potential to enhance the classifier's robustness against adversarial attacks %
(see \cref{sec:derandomization} for more discussion).
In this work, we show a generic procedure to derandomize any robust classifier, which preserves the extent of its robustness and certification. 

\paragraph{Our Contributions.}
\begin{itemize}

\item We show that bounded VC-dimension is sufficient for distributionally adversarial PAC-learning with a proper learning rule. [\Cref{sec:DRL},~\Cref{thm:realizable-sample-complexity,thm:agnostic-sample-complexity}]
\item One important application of our framework is to derive sample complexity bounds for randomized smoothing methods, unifying the line of work on robust learning in PAC-learning settings and randomized smoothing methods. [\Cref{sec:connect-RS-PAC}]

\item We show a general derandomization technique that preserves the extent of a randomized classifier's robustness and certification.
Additionally, our experimental findings suggest that this approach has the potential to enhance the robust accuracy of the initial classifier.
\end{itemize}

All missing proofs are in the Appendix. %

\paragraph{Further related work.} Prior work has studied the sample complexity of robust learning in the PAC-learning setting. Among others,~\citet{cullina2018pac} derived sample complexity results for this problem through a notion called adversarial VC-dimension.~\citet{attias2022improved}'s sample complexity depends on the number of perturbations allowed for
each instance.~\citet{montasser2019vc}'s upper bound depends on the VC and dual VC dimension of a hypothesis class, and they use an improper learning rule.~\citet{montasser2022adversarially} demonstrated a characterization of robust learning based on the one-inclusion graph of~\citet{haussler1994predicting}
adapted to robust learning. In our work unlike the prior work, we model the perturbation set as a set of distributions instead of a set of points and connect our notion to randomized smoothing methods.  There is also substantial work on learning from training data corrupted by malicious noise \cite{kearns1988learning,awasthi2017power,balcan2021noise}.

\section{Distributional Adversarial Loss}
\label{sec:DRL}

\subsection{Problem Setup}
\paragraph{Loss Function} %
We are given an instance space $\calX$ and label space $\calY=\{-1,+1\}$ and a distribution $\calD$ over $\calX\times \calY$. For each unperturbed (clean) input $(x,y)\sim \calD$, i.e. input $x$ with label $y$, the perturbation set $\calU(x)$ is a family of distributions, where a perturbation $z$ of $(x,y)$ is sampled from an adversarially chosen distribution $u\in\calU(x)$. Given that, the distributional adversarial loss of a classifier $h$ is defined as: %
\begin{align}
\label{definition:DAL}
\DR_{\calD}(h)=\Ex_{(x,y)\sim \calD} \insquare{ \max_{u\in \calU(x)}\insquare{\Ex_{z\sim u} \ind\insquare{h(z)\neq y}} } 
\end{align}
That is, for each non-adversarial example sampled from an underlying distribution $\calD$, the robust loss on $(x,y)$ is defined as the maximum expected loss over all the distributions in its perturbation set $\calU(x)$. The objective is to minimize the expected robust loss over all the samples coming from the underlying distribution $\calD$. 
First, we consider a setting where for each example $(x,y)$, the size of the perturbation set is bounded, i.e. $|\calU(x)|\leq k$ for some value $k>0$. Next, we consider two further extensions. 

In the first extension in~\Cref{sec:unbounded-distributions}, we investigate a setting where $\calU(x)$ is potentially unbounded, however, there exists a set of representative distributions $\calR(x)\subseteq \calU(x)$ where $|\calR(x)|\leq k$ and for each distribution $u\in \calU(x)\setminus \calR(x)$, there exists a representative distribution $r\in \calR(x)$ where the total variation distance between $u$ and $r$ is bounded. Consequently, in~\Cref{sec:connect-RS-PAC}, we demonstrate that this extension captures the sample complexity of randomized smoothing methods.

Furthermore, in~\Cref{sec:extension2}, we consider a scenario where the number of distribution perturbations is unbounded, however, there exists a set of \emph{representative} distributions of size at most $k$ such that each distribution $u\in \calU(x)$, is completely covered by $\calR(x)$, i.e. the probability density function of $u$ is point-wise bounded by the maximum of probability density functions of the distributions in $\calR(x)$.

\paragraph{PAC Learning of Distributional Adversarial Loss:} We study the \emph{sample complexity} for %
\emph{PAC-learning of distributional adversarial loss} in the realizable and agnostic settings. Given a hypothesis class $\calH\subseteq \calY^{\calX}$, our goal is to find a learning rule $\calL$ such that for any distribution $\calD$ over $\calX\times \calY$ finds a classifer $h\in \calH$ that competes with the predictor $h^*\in \calH$ where $h^*=\argmin_{h\in\calH} \DR_{\calD}(h)$, using a number of samples that is independent of the underlying distribution $\calD$. In the following, we formally define the notion of distributionally adversarial PAC learning
in realizable and agnostic settings.

\begin{definition}[Agnostic Distributionally Adversarial PAC learning]
A hypothesis class $\calH$ is agnostic distributionally adversarial PAC-learnable if there exists functions $n_{\calH}:(0,1)^2\rightarrow N$ and $m:(0,1)\rightarrow N$ and a learning algorithm $\calL$ with the following property: For every $\eps,\delta\in (0,1)$, and for every distribution $\calD$ over $\calX\times \calY$, when running the learning algorithm on a set $S=S_c\cup S_p$ where the set of clean (unperturbed) examples $S_c$ is consisting of $n\geq n_{\calH}(\eps,\delta)$ i.i.d. examples sampled from $\calD$ and for each example $(x,y)\in S_c$, $m\geq m(\eps)$ perturbations sampled from each $u\in \calU(x)$ and added to the perturbations set $S_p$, the algorithm returns $h\in \calH$ such that, with probability of at least $1-\delta$, 
\[\DR_{\calD}(h)\leq \min_{h^*\in \calH}\DR_{\calD}(h^*)+\eps\]
\end{definition}

Similarly, we can define Realizable Distributionally Adversarial PAC learning where the goal is to find a predictor $h\in \calH$ such that, with probability of at least $1-\delta$, $\DR_{\calD}(h)\leq \eps$.

\emph{We show that bounded VC-dimension is sufficient for distributionally adversarial PAC-learning.}~\Cref{thm:realizable-sample-complexity,thm:agnostic-sample-complexity} prove this in the realizable and agnostic case respectively.

\subsection{Learning Algorithm:} 
In this section, we describe our learning algorithm. Usually, in the PAC-learning setting, the idea is to sample enough number of examples from the underlying distribution, and find the best predictor $h^*$ on the sampled dataset. 
Furthermore, using uniform convergence guarantees, it can be argued that assuming enough number of examples are sampled from the underlying distribution, the performance of $h^*$ on the sampled dataset and the true distribution are close.

However, for our problem, we need a two-layer sampling procedure. This happens since in the definition of distributional adversarial loss (\Cref{definition:DAL}), there are two expectations, one for the underlying distribution $\calD$ over the clean (unperturbed) examples, and the second one for each perturbation distribution $u$ of an example coming from $\calD$. As a result, in order to derive generalization guarantees, first, we need to sample enough number of examples from the underlying distribution $\calD$. Furthermore, since each sampled example $(x,y)$ 
has a set of distributions $\calU(x)$
as its perturbation set, for each distribution $u\in \calU(x)$, we need to draw a number of samples from $u$ in order to argue about performance on each perturbation distribution $u$ in addition to the performance on the underlying distribution $\calD$. %

Our training procedure is as follows. First, we draw a sample set $S_c$ of clean (unperturbed) examples of size 
$n\geq n_{\calH}(\eps,\delta)$ i.i.d from $\calD$. Then, for each example $(x,y)\in S$, we draw %
$m\geq m(\eps)$ samples i.i.d. from each of the distributions $u\in \calU(x)$ (or from each $u\in \calR(x)$ when $\calU(x)$ is unbounded) and add to the perturbations set $S_p$ and let the training set be $S=S_c\cup S_p$. Therefore, $|S|\leq n\cdot m \cdot k+n$. For the training, we assume having access to an oracle $\DRERM$ that minimizes \emph{empirical} distributional adversarial loss:
\[\hat{h}\in \DRERM_{\calH}(S)=\argmin_{h\in \calH}\DR_{S}(h)\]
where the empirical distributional adversarial loss is defined as follows:
{\small\[\DR_{S}(h)=\frac{1}{n}\sum_{(x,y)\in S_{c}} \insquare{ \max_{u\in \calU(x)}\insquare{\frac{1}{m}\sum_{z\in \calU(x)\cap S_{p}} \ind\insquare{h(z)\neq y}}}\]}

Now we are ready to demonstrate our generalization guarantees for distributional adversarial loss. %

\subsection{Realizable Distributionally Adversarial PAC Learning:}
\label{section:realiable-PAC-DAL}

In this section, we focus on the setting where for each example $(x,y)$, the size of their perturbation set is bounded, i.e. $|\calU(x)|\leq k$.~\Cref{thm:realizable-sample-complexity} exhibits sample complexity bounds in the realizable setting. As mentioned before, we need a two-layer sampling procedure in our setting. Consequently,~\Cref{thm:realizable-sample-complexity} bounds the number of samples needed from the underlying distribution $\calD$ over unperturbed examples, and the number of perturbations needed to be sampled from each distribution $u$ in the perturbation set of a sampled example $x$.

\begin{theorem}[VC-dimension sample bound in the realizable distributionally adversarial case]
\label{thm:realizable-sample-complexity}
For any class $\calH$ and distribution $\calD$, a training sample $S_{c}$ of size 
$n=\calO\Big(\frac{1}{\eps}[VCdim(\calH)\log(\frac{mk}{\eps})+\log(\frac{1}{\delta})+\frac{1}{\eps}]\Big)$
where for each $(x,y)\in S_{c}$,  %
$m=\Omega(1/\eps^2\log(1/\eps))$
perturbations are sampled from each of the distributions $u\in\calU(x)$. Let $S_{p}$ denote the set of all perturbations, then $S=S_{c}\cup S_{p}$. Given sample set $S$, %
with probability $\geq 1-\delta$, every $h\in \calH$ with $\DR_{\calD}(h)\geq \eps$ has $\DR_{S}(h)>0$ (equivalently, every $h\in\calH$ with $\DR_S(h)=0$ has $\DR_{\calD}(h)<\eps$).
\label{thm:vc-dim-robustly-realizable}
\end{theorem}

In order to prove~\Cref{thm:vc-dim-robustly-realizable}, first we show~\Cref{lem:robust-loss} holds, which states the following:
Consider drawing a set $S$ of $n$ examples from $\calD$ where for each example $(x,y)\in S$, $m$ perturbations are drawn from each $u\in \calU(x)$ and are added to $S$. Let $A$ denote the event that there exists $h\in \calH$ with zero empirical distributional adversarial error on $S$ but true distributional adversarial error at least $\eps$. Now draw a \emph{fresh} test set $S'$ of $n$ examples from $\calD$ where for each example $(x,y)\in S'$, $m$ perturbations are drawn from each distribution $u\in \calU(x)$ and are added to $S'$. Let $B$ denote the event that there exists $h\in\calH$ with zero empirical distributional adversarial loss on $S$ but an empirical distributional adversarial loss at least $\eps/2$ on $S'$. We prove that $\Pr(B)\geq (2/5)\Pr(A)$. 

The purpose of this lemma is to show that we can argue about the error on a fresh test set instead of the true error on the underlying distribution. Later on, in~\Cref{thm:robust-loss-realizable-pre-Sauer}, we show that for large enough training and test sets, the error values on the training and test sets are close. This implies that the training error is also close to the true error on the underlying distribution, hence it cannot be the case that $\DR_S(h)=0$ but $\DR_{\calD}(h)>\eps$, and it proves~\Cref{thm:realizable-sample-complexity}.

\begin{lemma}
\label{lem:robust-loss}
Let $\calH$ be a concept class over a domain $\calX$. Let $S_{c}$ and $S'_{c}$ be
sets of $n$ clean (unperturbed) elements drawn from some distribution $\calD$ over $\calX$, where $n=\Omega(1/\eps^2)$. %
For each $(x,y)\in S_{c}$, %
$m=\Omega(1/\eps^2\log(1/\eps))$ perturbations sampled from each $u\in\calU(x)$ are added to a set $S_{p}$ and finally $S=S_{c}\cup S_{p}$. 
Similarly, $S'_{c}$ is augmented to get $S'$.
Let $A$ denote the event that there exists $h\in \calH$ with zero empirical distributional adversarial error on $S$ but true distributional adversarial error $\geq \eps$:
\begin{align*}
&\Ex_{(x,y)\sim \calD} \insquare{ \max_{u\in \calU(x)}\insquare{\Ex_{z\sim u} \ind\insquare{h(z)\neq y}} }\geq \eps ,\\
&\frac{1}{n}\sum_{(x,y)\in S_{c}} \insquare{ \max_{u\in \calU(x)}\insquare{\frac{1}{m}\sum_{z\in \calU(x)\cap S_{p}} \ind\insquare{h(z)\neq y}} }=0
\end{align*}
Let $B$ denote the event that there exists $h\in\calH$ with zero distributional adversarial loss on $S$ but distributional adversarial loss $\geq \eps/2$ on $S'$:
\begin{align*}
&\frac{1}{n}\sum_{(x,y)\in S'_{c}} \insquare{ \max_{u\in \calU(x)}\insquare{\frac{1}{m} \sum_{z\in \calU(x)\cap S'_{p}}\ind\insquare{h(z)\neq y}} }\geq \eps/2 ,\\
&\frac{1}{n}\sum_{(x,y)\in S_{c}} \insquare{ \max_{u\in \calU(x)}\insquare{\frac{1}{m}\sum_{z\in \calU(x)\cap S_{p}} \ind\insquare{h(z)\neq y}} }=0
\end{align*}
Then $\Pr(B)\geq (2/5)\Pr(A)$.
\end{lemma}

Next, we prove~\Cref{thm:robust-loss-realizable-pre-Sauer}. Proof of this theorem is similar to the original double-sampling trick by~\citet{vapnik1971uniform,blumer1989learnability} for showing sample complexity of PAC-learning. However, here, we also need to argue about the perturbations of the clean examples in the dataset. The idea is first to use the application of~\Cref{lem:robust-loss} to argue about the distributional adversarial loss on the test data instead of population distributional adversarial loss. Furthermore, for large enough training and test data, when sampled from the same distribution, it cannot be the case that the training distributional adversarial loss is low but the test distributional adversarial loss is large.
\begin{theorem}
\label{thm:robust-loss-realizable-pre-Sauer}
For any class $\calH$ and distribution $\calD$, a training sample $S_{c}$ of size $n\geq\frac{2}{\eps}[\log((5/2)\calH[2n\cdot m\cdot k])+\log(\frac{1}{\delta})+\frac{7}{\eps}]$
where for each $(x,y)\in S_{c}$, $m=\Omega(1/\eps^2\log(1/\eps))$ %
perturbations are sampled from each of the distributions $u\in\calU(x)$. %
Let $S_{p}$ denote the set of all perturbations, then $S=S_{c}\cup S_{p}$. Let $k=\max_{x\in \calX}|\calU(x)|$. 
Given sample set $S$, %
with probability $\geq 1-\delta$, every $h\in \calH$ with $\DR_{\calD}(h)\geq \eps$ has $\DR_{S}(h)>0$ (equivalently, every $h\in\calH$ with $\DR_S(h)=0$ has $\DR_{\calD}(h)<\eps$). Here, $\calH[.]$ is the growth function of $\calH$.
\end{theorem}

Putting together~\Cref{thm:robust-loss-realizable-pre-Sauer} and~\Cref{lem:robust-loss} and applying Sauer's Lemma proves~\Cref{thm:realizable-sample-complexity}. Details are deferred to the Appendix.

\subsection{Agnostic Distributionally Adversarial PAC Learning}
\label{section:agnostic-PAC-DAL}
In this section, we present our sample complexity bounds in the agnostic setting where for each clean example, their perturbation set has size at most $k$.
In this setting,~\Cref{thm:agnostic-sample-complexity} gives the sample complexity bound for %
distributionally adversarial PAC-learning. The idea to prove~\Cref{thm:agnostic-sample-complexity}, is similar to~\Cref{thm:realizable-sample-complexity}. First, we exhibit that it suffices to argue about the distributional adversarial loss on a fresh test data instead of the population distributional adversarial loss. Furthermore, when the test and training data are coming from the same distribution, it cannot be the case that the training distributional adversarial loss is low but the test distributional adversarial loss is large. Finally, by the application of Sauer's lemma the proof is complete. Details are deferred to Appendix.
\begin{theorem}[VC-dimension sample bound in the agnostic case]
\label{thm:agnostic-sample-complexity}
For any class $\calH$ and distribution $\calD$, a training sample $S_{c}$ of size $n=\calO\Big(\frac{1}{\eps^2}[VCdim(\calH)\log(\frac{mk}{\eps})+\log(\frac{1}{\delta})]\Big)$ where for each $(x,y)\in S_{c}$,  %
$m=\Omega(1/\eps^2\log(1/\eps))$ perturbations are sampled from each of the distributions $u\in\calU(x)$. Let $S_{p}$ denote the set of all perturbations, then $S=S_{c}\cup S_{p}$. Given sample set $S$, %
with probability $\geq 1-\delta$, for every $h\in \calH$, $|\DR_{\calD}(h)-\DR_{S}(h)|\leq \eps$.
\label{thm:vc-dim-robustly-agnostic}
\end{theorem}

\subsection{Extension to an arbitrary number of distributions}
\label{sec:unbounded-distributions}

\subsubsection{Model I}
\label{section:extension-1}
In this section, we consider the scenario where the number of distribution perturbations is unbounded, however, there exists a set $\calR(x)$ of \emph{representative} distributions of size at most $k$ such that for each distribution $u\in \calU(x)$, there exists a distribution $u_0\in \calR(x)$ such that the variation distance between $u$ and $u_0$ is bounded (at most $\eps'$). 
We show that using a similar learning algorithm that was used in the previous section, with the main difference that after drawing $n\geq n_{\calH}(\eps,\delta)$ i.i.d from $\calD$ and adding to the training set, for each clean example $(x,y)\in S$, the perturbations are drawn from the representative distribution sets $\calR(x)$ instead of the true distribution set $\calU(x)$. %
In~\Cref{thm:vc-dim-robustly-realizable-model-representative},
we prove that training a predictor that minimizes the empirical distributional adversarial loss on $S$, will also minimize the population distributional adversarial loss with respect to the true perturbation sets $\calU(.)$. A similar sample complexity bound holds in the agnostic setting (\Cref{thm:vc-dim-robustly-agnostic-model-representative}).
We show the connection of this model to randomized smoothing methods in~\Cref{sec:connect-RS-PAC}.

\begin{theorem}[realizable case]
For any class $\calH$ and distribution $\calD$, a training sample $S_{c}$ of size $n=\calO\Big(\frac{1}{\eps}[VCdim(\calH)\log(\frac{mk}{\eps})+\log(\frac{1}{\delta})+\frac{1}{\eps}]\Big)$, where for each $(x,y)\in S_{c}$,  
$m=\Omega(1/\eps^2\log(1/\eps))$ perturbations are sampled from each of the distributions $u\in\calR(x)$. Let $S_{p}$ denote the set of all perturbations, then $S=S_{c}\cup S_{p}$. Given sample set $S$, %
with probability $\geq 1-\delta$, %
every $h\in\calH$ with $\DR_S(h)=0$ has $\DR_{\calD}(h)<\eps+\eps'$. $\eps'$ is an upper bound on the total variation distance between each distribution $u\in \calU(x)$ and the closest representative distribution $u_0\in \calR(x)$.
\label{thm:vc-dim-robustly-realizable-model-representative}
\end{theorem}

\subsubsection{Connection to Randomized Smoothing}
\label{sec:connect-RS-PAC}
Randomized smoothing constructs a new smoothed classifier $g$ from a base classifier $f$. Given an input $x$, classifier $g$ returns the class that the base classifier $f$ is most likely to return given $x$ is perturbed using a Gaussian noise:
\[g(x)=\argmax_{c\in \calY}\Pr(f(x+\eta)=c)\]
where $\eta\sim \calN(0,\sigma^2I)$. %
~\citet{cohen2019certified} exhibit certified robustness guarantees that a smoothed classifier $g$ is robust around a clean input $x$ within certain $\ell_2$ radius. They show when the base classifier $f$ classifies $\calN(x,\sigma^2I)$, the most probable class $c_A$ is returned with probability $p_A$, and the runner-up class is returned with probability $p_B$, then $g(x+\gamma)=c_A$ for all $\|\gamma\|_2<R$, where $R=\sigma/2(\Phi^{-1}(p_A)-\Phi^{-1}(p_B))$ and $\Phi^{-1}$ is the inverse of the standard Gaussian CDF. %

Our result is orthogonal to the guarantee exhibited by~\citet{cohen2019certified}. In their framework, the goal is that the smoothed classifier outputs the same prediction under perturbation as the base classifier, whether it is a correct prediction or not. However, in our framework, the goal is to derive sample complexity results such that assuming the unperturbed points are coming from a distribution, we train a classifier $f$ such that after adding Gaussian noise to a clean input, it still predicts the correct label with high probability. Furthermore, our guarantees extend to the smoothed classifier $g$.

In order to exhibit sample complexity guarantees for randomized smoothing, we use~\Cref{thm:vc-dim-robustly-realizable-model-representative,thm:vc-dim-robustly-agnostic-model-representative} where $k=1$. %
For each example $(x,y)\sim \calD$, the representative distribution set $\calR(x)$ consists of one single distribution that is a  Gaussian around $x$, i.e.  $\calN(x,\sigma^2I)$. During the test time, each example $x$ can be perturbed by adding a limited perturbation $\gamma$ to get $x'$.
Given $x'$ in the test time, the classifier $g$ adds a Gaussian noise $\eta\sim \calN(0,\sigma^2I)$ to $x'$ and outputs $g(x')=\argmax_{c\in \calY}\Pr(f(x'+\eta)=c)$. Considering our model, the true perturbation set $\calU(x)$ for a clean input $x$ is a Gaussian around $x'$, i.e.  $\calN(x',\sigma^2I)$, for $\|x-x'\|\leq \gamma$. 
Given $\|x-x'\|\leq \gamma$, the total variation distance between the Gaussians around $x$ and $x'$ is bounded and is a function of $\gamma$ and $\sigma$, i.e. $d(\gamma,\sigma)$\footnote{\citet{devroye2018total}(Proposition 2.2.) derive an upper bound on the total variation distance of two multivariate Gaussians. }. %
Now by~\Cref{thm:vc-dim-robustly-realizable-model-representative,thm:vc-dim-robustly-agnostic-model-representative}, it suffices to do training on the perturbations coming from the representative distributions $\calR(.)$ in order to get guarantees with respect to the true perturbation sets $\calU(.)$. 

Consequently, by~\Cref{thm:vc-dim-robustly-agnostic-model-representative},  for any class $\calH$ and distribution $\calD$, we get a training sample $S_{c}$ of size $n=\calO\Big(\frac{1}{\eps^2}[VCdim(\calH)\log(\frac{mk}{\eps})+\log(\frac{1}{\delta})]\Big)$, where for each $(x,y)\in S_{c}$,  
$m=\Omega(1/\eps^2\log(1/\eps))$ perturbations are sampled from each of the distributions $u\in\calR(x)$. Let $S_{p}$ denote the set of all perturbations, then $S=S_{c}\cup S_{p}$. Given sample set $S$, with probability $\geq 1-\delta$, for every $h\in \calH$, $\abs{\DR_{\calD}(h)-\DR_{S}(h)}\leq d(\gamma,\sigma)+\eps$. Where $\DR_{\calD}$ is defined with respect to the true perturbation sets $\calU(.)$. This means, with probability at least $1-\delta$,
\begin{align*}
&\DR_{\calD}(h)\\
=&\Ex_{(x,y)\sim \calD}\Big[\max_{x':\|x'-x\|\leq \gamma}\Ex_{\eta\sim \calN(0,\sigma^2I)}[h(x'+\eta)\neq y]\Big]\\
&\qquad\qquad\leq \DR_{S}(h)+d(\gamma,\sigma)+\eps 
\end{align*}
which is equivalent to saying with high probability, the expected adversarial error over all the examples coming from $\calD$ is bounded.

We can also use our framework to defend against multiple adversarial attacks. %
Consider an adversary that first alters the brightness of an image by scaling all its pixel values uniformly and then applies a bounded $\ell_2$ perturbation. Suppose the brightness level varies between $0$ and $U$.  For each unperturbed input, we first adjust the brightness levels to $\{U/k, 2U/k, \ldots, U\}$ uniformly and then add Gaussian noise to each modified image. Now, by using $k$ distributions for each unperturbed input in this manner, we can achieve robustness against multiple attacks.

\subsubsection{Model II}
\label{sec:extension2}
In this section, we consider the scenario where the number of distribution perturbations is unbounded, however, there exists a set of \emph{representative} distributions of size at most $k$ such that each distribution $u\in \calU(x)$, is completely covered by $\calR(x)$, i.e. the probability density function of $u$ is point-wise bounded by the maximum of probability density functions of the distributions in $\calR(x)$. %
Here, the sampling procedure is similar to the one used in~\Cref{section:extension-1}.\Cref{thm:vc-dim-robustly-realizable-extension-2,thm:vc-dim-robustly-agnostic-extension-2} exhibit the generalization guarantees in the realizable and agnostic settings.

\begin{theorem}[realizable case]%
For any class $\calH$ and distribution $\calD$, a training sample $S_{c}$ of size $n=\calO\Big(\frac{1}{\eps}[VCdim(\calH)\log(\frac{mk}{\eps})+\log(\frac{1}{\delta})+\frac{1}{\eps}]\Big)$, where for each $(x,y)\in S_{c}$, $m=\Omega(1/\eps^2\log(1/\eps))$ perturbations are sampled from each of the distributions $u\in\calR(x)$. Let $S_{p}$ denote the set of all perturbations, then $S=S_{c}\cup S_{p}$. Given sample set $S$, %
with probability $\geq 1-\delta$, %
every $h\in\calH$ with $\DR_S(h)=0$ has $\DR_{\calD}(h)<k\eps$.
\label{thm:vc-dim-robustly-realizable-extension-2}
\end{theorem}

\section{Derandomization}
\label{sec:derandomization}
The classifiers discussed above add noise to the perturbed input to achieve robustness and hence are randomized. We modeled this phenomenon as the adversary getting to pick from a set of distributions over perturbed inputs, namely $\calU(x)$, instead of a single perturbed input. Below we show how such classifiers can be derandomized in a general fashion while retaining the performance of the original classifier. 

We derandomize the part where the classifiers add noise to the input. Thus, we model the adversary in the traditional sense, i.e., for the clean input $x$ the adversary picks a perturbed input $x'$ from an allowed set of perturbations $\calA(x)$. Our classifier $h$ takes the input $x'$ and uses randomness $R$ sampled according to $\calR$ to make a prediction $h(x',R)$. Further, we also show how to derandomize a randomized certification procedure $\rho(x',R)$. The parameter $\rho(x',R)$ guarantees that all inputs within distance $\rho(x',R)$ from $x'$, receive the same label as ${x'}$. As mentioned in the introduction, there are a few advantages to such a derandomization. First, in critical decisions, we desire that the output of our classifier be deterministic so that it always labels a given input with the same label (same for the certification radius). Second, such a derandomization could potentially be useful in boosting the robust accuracy of the classifier against various adversaries (more on this is discussed below).  

The procedure is simple to describe: given the randomized classifier $h$ which on input $x'$ makes the prediction $h(x',R)$ (where $R$ is the randomness sampled according to some distribution $\mathcal{R}$), we pre-sample multiple copies of the randomness needed during inference, say $R_1,R_2,\ldots,R_t$ and define a new \emph{deterministic} classifier $h^{(R_1,R_2,\ldots, R_t)}$ as $\plural(h(x',R_i)\mid i\in [t])$. We show that when $t=\Omega(\log|\calA|)$ then with high probability over the choice of $R_1,R_2,\ldots,R_t$ the deterministic classifier $h^{(R_1,R_2,\ldots, R_t)}$ has an adversarial loss related closely to the performance of the original classifier, i.e.,  $\DR_{\calD}(h)$. %
Recall that for the purposes of derandmization we are modeling the adversary in the traditional sense, i.e., for each clean input $x$ the adversary is allowed to choose a perturbed input $x'$ from an allowed set $\calA(x)$. Hence, for a classifier $h$ the definition of $\DR_{\calD}(h)$ is modified accordingly as 
\[
\Ex\limits_{(x,y)\sim \calD}\left[ \max_{x'\in \calA(x)}\insquare{\Ex\limits_{R\sim \calR} \ind\insquare{h(x',R)\neq y}}\right].
\]

\begin{theorem}[Derandomizing a randomized robust classifier]
\label{thm:derandomization_classifier}
Suppose $h$ is a randomized classifier which on input $x$ uses randomness $R\sim \mathcal{R}$ and outputs a label in $\mathcal{Y}$.
Let $\calA(x)$ be the set of perturbed versions of $x$ from which the adversary chooses and define
$
\epsilon(x,y) \coloneqq \max_{x'\in \calA(x)}\insquare{\Ex\limits_{R\sim \calR} \ind\insquare{h(x',R)\neq y}}
$.
Also, suppose that $\DR_{\calD}(h) \leq \epsilon$, i.e., $\Ex_{(x,y)\sim \calD}[\epsilon(x,y)]  \leq \epsilon$, for some $\epsilon>0$. 
For any $0<\eta<1/2$, let $\epsilon(\eta) \coloneqq \Pr_{(x,y)\sim \calD}[\epsilon(x,y)\geq1/2-\eta]$ (notice that $\epsilon(\eta)\leq \frac{2\epsilon}{1-2\eta}$). Let $\delta>0$ be a parameter. Set $t=\frac{100}{\eta^2}\cdot \log(|\calA(x)|/\delta)$ and define $h^{(R_1,R_2,\ldots,R_t)}(x') \coloneqq \plural(h(x',R_i)\mid i\in [t])$. 
Then, with probability at least $1-\delta$ over the choice of $R_1,R_2,\ldots R_t$ sampled iid according to $\calR$ we have
\[
\DR_{\calD}(h^{(R_1,R_2,\ldots,R_t)}) \leq \delta + \epsilon(\eta), 
\]
where 
\begin{align*}
&\DR_{\calD}(h^{(R_1,R_2,\ldots,R_t)})\\  
=& \Pr_{(x,y)\sim \calD}[\exists x'\in \calA(x)\colon h^{(R_1,R_2,\ldots,R_t)}(x')\neq y ].
\end{align*}
\end{theorem}

\emph{Boosting accuracy via Derandomization:} 
Given a randomized classifier $h$, the procedure detailed in~\Cref{thm:derandomization_classifier}, fixes  
the randomness used during inference beforehand. However, during training, we do not modify $h$ according to this pre-fixed randomness. It is natural to wonder %
if this knowledge of the pre-fixed randomness can be used during training to help boost the accuracy. We show in \cref{sec:experiments} that this is indeed the case empirically, by training our classifiers according to the pre-fixed randomness to be used during inference. An interesting question that arises here is whether under suitable assumptions a variant of \cref{thm:derandomization_classifier} can be proven to capture the above phenomenon. Currently, \cref{thm:derandomization_classifier} prescribes taking a majority vote over multiple instantiations of the original classifier $h$, whereas we would like to be able to train a classifier based on the pre-fixed randomness which might result in a classifier not corresponding to any instantiations of $h$.%

Next we show that it is also possible to derandomize a randomized certification procedure. Suppose we have a classifier $h$ and a randomized certification mechanism $\rho$ which on input $x'$ uses randomness $R$ (sampled according to $\mathcal{R}$) and outputs a certifiable radius $\rho(x',R)$.
We define $\rho^{(R_1,R_2,\ldots, R_t)}$ as the median of $\rho(R_i)$'s, i.e., $\median(\rho(R_i)\mid i\in [t])$ and this serves as our derandomized proxy for the certifiable radius. We show that the behavior of the median is statistically close to the behavior of $\rho(R)$. By this we mean the following: for any input $x'$ let $\robreg(h,x')$ denote a parameter which serves as the measure of the region around $x'$ where the label according to the classifier $h$ doesn't change. For instance, this could be the radius of the $\ell_p$ ball around $x'$ in which the label according to $h$ doesn't change. 
Our certification procedure $\rho$ serves as an approximation to $\robreg(h,x')$ and it is usually desired that with high probability $\rho(x',R)\in [(1-\beta) \robreg(h,x'),(1+\alpha)\robreg(h,x')]$ for some values of $\beta>0$ and $\alpha>0$ as close to $0$ as possible.
We are interested in preserving the property that $\rho(x',R)\in [(1-\beta) \robreg(h,x'),(1+\alpha)\robreg(h,x')]$.

\begin{theorem}[Derandomizing a certifiably robust classifier]
\label{thm:derandomization_certifiable}
Let $h$ be a hypothesis and $\rho$ be a certification procedure which on input $x'$ uses randomness $R\sim \mathcal{R}$ and outputs $\rho(x',R)\in \R_{\geq 0}$. Further, let $\robreg(h,x')$ be as defined above and let $\gamma(\rho,x,y)$ be defined as 
\[
\max_{x'\in \calA(x)}\insquare{\Ex\limits_{R\sim \calR} \ind\insquare{\frac{\rho(x',R)}{\robreg(h,x')}\notin [1-\beta,1+\alpha]}}
\]
for some positive parameters $\alpha, \text{ and }\beta$. 
For any $0<\eta<1/2$ define $\epsilon(\eta)\coloneqq  \Pr_{(x,y)\sim \calD} \insquare{\gamma(\rho,x,y)\geq 1/2-\eta}$. 
Set $t=\frac{100}{\eta^2}\cdot \log(|\calA(x)|/\delta)$ and define $\rho^{(R_1,R_2,\ldots,R_t)}(x') \coloneqq \median(h(x',R_i)\mid i\in [t])$. Let $\gamma(\rho^{(R_1,R_2,\ldots,R_t)},x,y)$ be defined as
\begin{align*}
    \max_{x'\in \calA(x)} \insquare{\ind \insquare{ \frac{\rho^{(R_1,R_2,\ldots,R_t)}(x')}{\robreg(h,x')}\notin [1-\beta,1+\alpha]}}.    
\end{align*}

Then, with probability $1-\delta$ over the choice of $R_1,R_2,\ldots R_t$ sampled iid according to $\calR$ we have
\begin{align*}
    \Ex_{(x,y)\sim \calD} [ \gamma(\rho^{(R_1,R_2,\ldots,R_t)},x,y)]  \leq \epsilon(\eta) + \delta. 
\end{align*}
\end{theorem}

\section{Experiments}
\label{sec:experiments}

To show the effectiveness of our derandomization method in practice, we implement our framework on the method proposed by~\citet{dong2023adversarial}. They use random projection filters to improve the adversarial robustness. 
In their method, they replace some of the CNN layers with random filters which are non-trainable, both during training and inference, and by doing so take away some of the adversarial impact on the input. More precisely, they re-instantiate the random filters for each training example and training is done only on the remaining weight parameters of the CNN.
Further, during inference time fresh random filters are used.
In contrast, in our method, we fix the random filters to be used (in the style of \cref{thm:derandomization_classifier} %
) and train a few models with the pre-fixed convolutional filters. This fixing is done randomly at the beginning of the training. During the inference time, we use the majority vote of these trained models on the test input. We follow closely the experimental setup of \cite{dong2023adversarial} and refer the reader to their paper for further details . Below we outline the setup and our modification to implement the derandomization approach. 
Codes are publicly available at \href{https://github.com/TTIC-Adversarial-Robustness/multiple_rpf}{{https://github.com/TTIC-Adversarial-Robustness/multiple\_rpf}}.

\subsection{Experimental Setup}
\paragraph{Datases, Models and Computational Resources}
We use the CIFAR-10 dataset for our experiments that has 10 categories and contains $60K$ colored images with size $32\times 32$, including $50$K training images and $10$K validation. We use the ResNet-$18$ architecture model on CIFAR-$10$. The random filters are applied at the first layer where the weights of the random filters are sampled as iid $N(0,1/r^2)$ ($r$ is the Kernel size). The ratio of the random filters to the total number of filters is kept at $0.75$ (this parameter is denoted as $N_r/N$ in \cite{dong2023adversarial}). %
For our experiments, we utilized a %
computing node with 24 cores, 192 GB of system RAM, and 4 NVIDIA GeForce RTX 2080 Ti GPUs, each with 11 GB of dedicated VRAM. %

\paragraph{Training Strategy}
As we have described earlier, we have multiple models corresponding to each instantiation of the random coins tossed in the beginning. %
For training a \emph{single} model, as in \cite{dong2023adversarial} we adhere to the established protocol of a state-of-the-art adversarial training strategy for setting up our experiments on CIFAR-$10$. The network undergoes training for $200$ epochs with a batch size of $128$ using SGD with a momentum of $0.9$. The learning rate is fixed at $0.1$, and the weight decay is set to $5 \times 10^{-4}$.
Employing a piecewise decay learning rate scheduler, we initiate a decay factor of $0.1$ at the $100^{th}$ and $150^{th}$ epochs.
For generating adversarial examples, we utilize PGD-10 with a maximum perturbation size $\eps = \frac{8}{255}$, and the step size of PGD is specified as $\frac{2}{255}$.
Although, the training phase only considers PGD adversarial examples, yet during evaluation we consider other attacks.

\paragraph{Attacks}
To assess the adversarial robustness robustness of our algorithm, we utilize Projected Gradient Descent (PGD) \cite{MadryMSTV18}, Fast Gradient Sign Method (FGSM) \cite{szegedy2013intriguing} and Auto Attack \cite{Croce020a}. Following the standard protocol for attack configuration, we set the maximum perturbation size $\varepsilon$ to $8/255$ for PGD, FGSM and Auto Attack. For PGD, the step size is established at $2/255$ over $20$ steps.

\paragraph{Benchmark}
We compare various versions of our derandomized majority predictor model where the majority votes are taken over $n$ base models each initialized with randomly chosen non-trainiable filters, with $n=1,2,\ldots,14$. 
We also include a comparison with a variant of the model of ~\citet{dong2023adversarial} where a majority vote is taken over multiple inferences performed (once, $10$, $20$ and $30$ times) on a single test input using fresh randomness each time (such a model with a majority over $5$ votes was considered in \cite{dong2023adversarial}). This helps us better contrast the accuracy gains due to derandomization.

\paragraph{Results}
We can consistently match and outperform the benchmark by some percentage points in accuracy, even when the majority vote is taken over $30$ trials for the benchmark, by just using $11$ models with pre-fixed random filters.
We tabulate the results in \cref{tab:model_performance_comparison} below. Further plots detailing the results for varying number of trials can be found in the appendix.

\begin{table}[h]
\centering
\begin{tabular}{|l|l|l|}
\hline
Metric & Our Model &\citet{dong2023adversarial} \\
\hline
Natural Acc     & 0.8603                                                  & 0.8580                                                  \\ \hline
PGD Acc         & 0.6302                                                  & 0.6249                                                  \\ \hline
FGSM Acc        & 0.6600                                                  & 0.6452                                                  \\ \hline
Auto Acc        & 0.6696                                                  & 0.6591                                                  \\ \hline
\end{tabular}
\caption{Comparison of Model Performances. The middle column corresponds to our model where we take the majority vote over $11$ pre-fixed random filters. Second column corresponds to the model used by~\citet{dong2023adversarial} where they take the majority vote over $30$ fresh random filters.}
\label{tab:model_performance_comparison}
\end{table}

\paragraph{Conclusion}
We studied %
the new notion of distributional adversarial loss and proved generalization guarantees for it, and showed how it derives sample complexity bounds for randomized smoothing methods.
Furthermore, we show a general derandomization technique which preserves the extent of a randomized classifier's robustness and certification. In terms of further directions, as highlighted in Section \ref{sec:derandomization}, it is intriguing to explore the impact of employing pre-fixed randomness in the derandomization process, during training. Does this contribute to enhanced robustness? Our experiments, which incorporate random projection filters, provide some supporting evidence for this notion. %

\subsection*{Acknowledgements}
This work was supported in part by the National Science Foundation under grants CCF-2212968 and ECCS-2216899, by the Simons Foundation under the Simons Collaboration on the Theory of Algorithmic Fairness, and by the Defense Advanced Research Projects Agency under cooperative agreement HR00112020003, and by the Office of Naval Research MURI Grant N000142412742. The views expressed in this work do not necessarily reflect the position or the policy of the Government and no official endorsement should be inferred. Most of the research was done when Chen Dan was a postdoctoral researcher at TTIC and Prabhav Jain was an intern there.  We thank Anand Bhattad for helpful discussions. 
\bibliography{ref}
\bibliographystyle{abbrvnat}
\newpage
\appendix
\onecolumn
\section*{Checklist}

 \begin{enumerate}

 \item For all models and algorithms presented, check if you include:
 \begin{enumerate}
   \item A clear description of the mathematical setting, assumptions, algorithm, and/or model. [Yes, Section 2]
   \item An analysis of the properties and complexity (time, space, sample size) of any algorithm. [Yes, we provide sample complexity results for our objective in the PAC-learning setting.]
   \item (Optional) Anonymized source code, with specification of all dependencies, including external libraries. [Yes]
 \end{enumerate}

 \item For any theoretical claim, check if you include:
 \begin{enumerate}
   \item Statements of the full set of assumptions of all theoretical results. [Yes]
   \item Complete proofs of all theoretical results. [Yes, Appendix D]
   \item Clear explanations of any assumptions. [Yes]     
 \end{enumerate}

 \item For all figures and tables that present empirical results, check if you include:
 \begin{enumerate}
   \item The code, data, and instructions needed to reproduce the main experimental results (either in the supplemental material or as a URL). [Yes]
   \item All the training details (e.g., data splits, hyperparameters, how they were chosen). [Yes, in the experiments section]
         \item A clear definition of the specific measure or statistics and error bars (e.g., with respect to the random seed after running experiments multiple times). [Yes, in Table 1, we report accuracies that are majority-vote of multiple runs of a model.]
         \item A description of the computing infrastructure used. (e.g., type of GPUs, internal cluster, or cloud provider). [Yes, Section 4.1, paragraph 1.]
 \end{enumerate}

 \item If you are using existing assets (e.g., code, data, models) or curating/releasing new assets, check if you include:
 \begin{enumerate}
   \item Citations of the creator If your work uses existing assets. [Yes, Section 4]
   \item The license information of the assets, if applicable. [Not Applicable]
   \item New assets either in the supplemental material or as a URL, if applicable. [Yes, supplemental material]
   \item Information about consent from data providers/curators. [Not Applicable]
   \item Discussion of sensible content if applicable, e.g., personally identifiable information or offensive content. [Not Applicable]
 \end{enumerate}

 \item If you used crowdsourcing or conducted research with human subjects, check if you include:
 \begin{enumerate}
   \item The full text of instructions given to participants and screenshots. [Not Applicable]
   \item Descriptions of potential participant risks, with links to Institutional Review Board (IRB) approvals if applicable. [Not Applicable]
   \item The estimated hourly wage paid to participants and the total amount spent on participant compensation. [Not Applicable]
 \end{enumerate}

 \end{enumerate}
\section{Potential Broader Impact}
\label{section:broader-impact}
Studying robustness to adversarial attacks is crucial since there is a threat that adversarial attacks are deployed in the real world.
Making machine learning technologies resistant to adversarial attacks makes machine learning models more reliable to be deployed in areas such as automated driving and healthcare. 
The methods presented in this paper further our understanding of various mechanisms which enhance the robustness of classifiers.

\section{Further Details of Experimental Results}

\begin{figure*}[ht]
  \centering
  \subfigure[Natural Accuracy]{
    \includegraphics[width=0.3\textwidth]{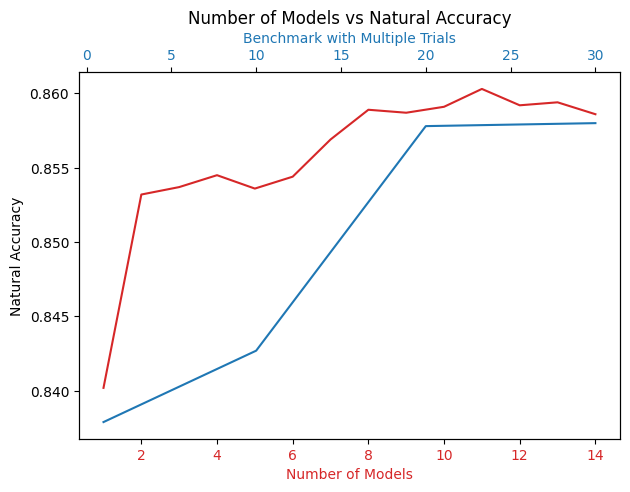}
    \label{fig:natural_acc}
  }
  \subfigure[FGSM]{
    \includegraphics[width=0.3\textwidth]{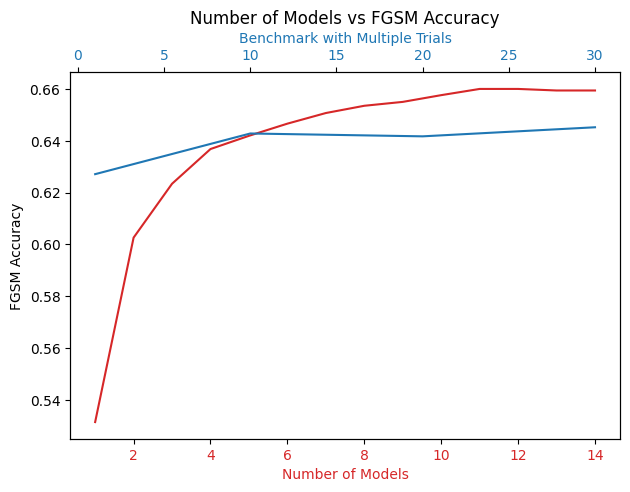}
    \label{fig:fgsm_acc}
  }

  \subfigure[PGD]{
    \includegraphics[width=0.3\textwidth]{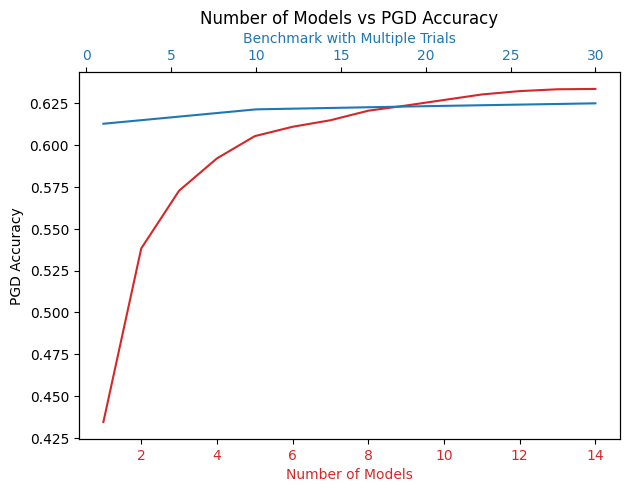}
    \label{fig:pgd_acc}
  }
  \subfigure[Auto Attack]{
    \includegraphics[width=0.3\textwidth]{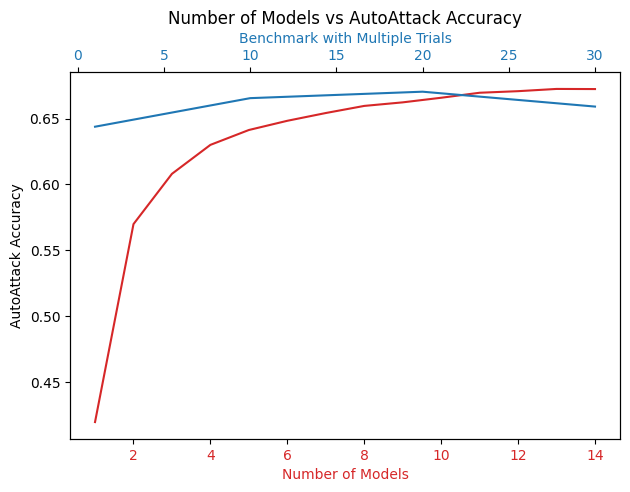}
    \label{fig:auto_attack_acc}
  }
    \subfigure[Bar graph comparing our model against the benchmark for various adversaries: blocks $1$, $3$ and $5$ correspond to the number of derandomized models used, where as blocks $2$ and $5$ correspond to the number of trails by the benchmark model ]{\includegraphics[width=0.45\textwidth]{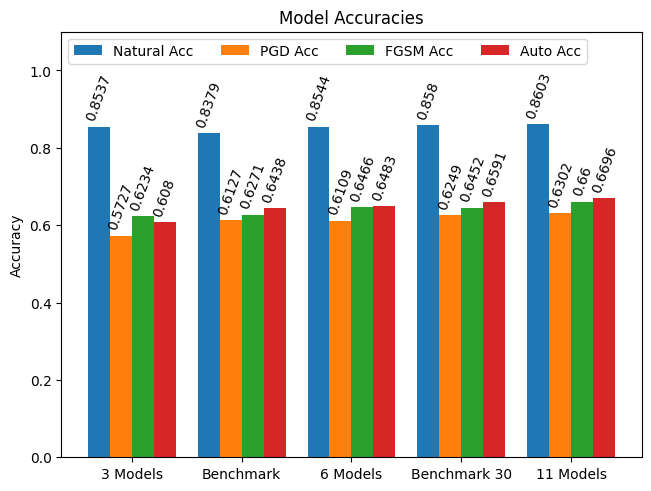}   \label{fig:histogram}}
 
  \caption{Results of derandomizing the Random Projection Filters framework of \cite{dong2023adversarial} using \cref{thm:derandomization_classifier}. The red and blue curves correspond to the derandomized model and the benchmark respectively.}
  \label{fig:our_model_vs_benchamark}
\end{figure*}

In \cref{fig:our_model_vs_benchamark} the blue curve corresponds to the benchmark and the red curve corresponds to our model.  The \cref{fig:natural_acc,fig:fgsm_acc,fig:pgd_acc,fig:auto_attack_acc} denote the performance for various adversaries, namely, no adversary (natural accuracy), FGSM, PGD and Auto Attack. %
We also collate the results in a bar graph, \cref{fig:histogram}.

\section{Missing Proofs}
\subsection{Proof of~\Cref{lem:robust-loss}}
\begin{proof}
For each input $(x,y)$, distribution $u\in \calU(x)$ and perturbation $z\sim u\in\calU(x)$, define the random variable %
$Z_{(x,y,u,z)} =\ind\insquare{h(z)\neq y}$. %
Let $Z_{(x,y,u)} = \sum_{z\in u\cap S'_{perturb}}\ind [h(z)\neq y]$. Let $\hat{Z}_{(x,y,u)} = \Ex_{z\sim u}\ind [h(z)\neq y]$.
By Hoeffding bound, for each example $(x,y)$ and $u\in \calU(x)$,
\begin{align*}
&\Pr\Bigg[\abs{\frac{1}{m}Z_{(x,y,u)}-\hat{Z}_{(x,y,u)}}\geq \eps/8\Bigg] 
= \Pr\Bigg[\abs{Z_{(x,y,u)}-m\cdot\hat{Z}_{(x,y,u)}}
\geq \frac{m\eps}{8}\Bigg]\leq 2e^{-2(m\eps/8)^2/m} = 2e^{-\frac{m\eps^2}{32}}
\end{align*}

Therefore, for a fixed example $(x,y)$,
\begin{align}
\label{eqn:hoeffding-3}
&\Pr\Bigg[\max_{u\in\calU(x)}\abs{\frac{1}{m}Z_{(x,y,u)}-\hat{Z}_{(x,y,u)}}\geq \eps/8\Bigg] \leq 2e^{-\frac{m\eps^2}{32}}
\end{align}
which implies that:%
\[\Pr\Bigg[\abs{\max_{u\in\calU(x)}\frac{1}{m}Z_{(x,y,u)}-\max_{u\in\calU(x)}\hat{Z}_{(x,y,u)}}\geq \eps/8\Bigg] \leq 2e^{-\frac{m\eps^2}{32}}\]

Therefore, for each input $(x,y)$, given $m\geq 32/\eps^2 \log(200/\eps)$:
\begin{align}
\label{eqn:hoeffding-10}
\Pr_{r\sim \calR}
\Bigg[\abs{\max_{u\in\calU(x)}\frac{1}{m}\sum_{z\in u\cap S'_{perturb}}\ind [h(z)\neq y]- \max_{u\in\calU(x)}\Ex_{z\sim u}\ind [h(z)\neq y]}\geq  \eps/8\Bigg] \leq \eps/100
\end{align}

where $r\sim \calR$ is the randomness used for generating $mk$ perturbations for each $(x,y)\in \text{supp}(\calD)$.

Let $Y(x,y)$ be an indicator random variable corresponding to input $(x,y)$ defined as follows:
\[Y(x,y)=\ind\Bigg[\abs{\max_{u\in\calU(x)}\frac{1}{m}\sum_{z\in u\cap S'_{perturb}}\ind [h(z)\neq y]- \max_{u\in\calU(x)}\Ex_{z\sim u}\ind [h(z)\neq y]}> \eps/8\Bigg]\]

Then:
\begin{align*}
&\Ex_{(x,y)\sim \calD}%
\Ex_{r\sim\calR}\abs{\max_{u\in\calU(x)}\frac{1}{m}\sum_{z\in u\cap S'_{perturb}}\ind [h(z)\neq y]- \max_{u\in\calU(x)}\Ex_{z\sim u}\ind [h(z)\neq y]}\\
\leq&\Ex_{(x,y)\sim \calD} \Big[\Pr[Y(x,y)=1](1)+ \Pr[Y(x,y)=0](\eps/8)\Big]=\eps/100+\eps/8=27\eps/200
\end{align*}

Therefore, by Markov's inequality: %
\begin{align}
\label{eqn:hoeffding-5}
\Pr_{r\sim\calR}
\Bigg[\abs{\Ex_{(x,y)\sim \calD}\Big[\max_{u\in\calU(x)}\frac{1}{m}\sum_{z\in u\cap S'_{perturb}}\ind [h(z)\neq y]\Big]-\Ex_{(x,y)\sim \calD}\Big[\max_{u\in\calU(x)}\Ex_{z\sim u}\ind [h(z)\neq y]\Big]}\geq \eps/4\Bigg] \leq 54/100
\end{align}

Now, for each input $(x,y)$, define the random variable $f(x,y)=\max_{u\in\calU(x)}\frac{1}{m}\sum_{z\in u\cap S'_{perturb}}\ind [h(z)\neq y]$. By Hoeffding bound:
\begin{align}
\label{eqn:hoeffding-4}
&\Pr_{S'_{clean}\sim \calD^n}\Bigg[\abs{\frac{1}{n}\sum_{(x,y)\in S'_{clean}}f(x,y) - \Ex_{(x,y)\sim D} f(x,y)}\geq \eps/4\Bigg]\leq 2e^{-\frac{n\eps^2}{8}}
\end{align}

Therefore, given $n\geq 13/\eps^2\geq \frac{8}{\eps^2} \log(200/6)$:
\begin{align}
\label{eqn:hoeffding-6}
&\Pr_{S'_{clean}\sim \calD^n}\Bigg[\abs{\frac{1}{n}\sum_{(x,y)\in S'_{clean}}\max_{u\in\calU(x)}\frac{1}{m}\sum_{z\in u\cap S'_{perturb}}\ind [h(z)\neq y] - \Ex_{(x,y)\sim D} \Big[\max_{u\in\calU(x)}\frac{1}{m}\sum_{z\in u\cap S'_{perturb}}\ind [h(z)\neq y]\Big]}\geq \eps/4\Bigg]\nonumber\\
&\leq 6/100
\end{align}

Now, we define bad events 
\[\calB_1:\abs{\Ex_{(x,y)\sim \calD}\Big[\max_{u\in\calU(x)}\frac{1}{m}\sum_{z\in u\cap S'_{perturb}}\ind [h(z)\neq y]\Big]-\Ex_{(x,y)\sim \calD}\Big[\max_{u\in\calU(x)}\Ex_{z\sim u}\ind [h(z)\neq y]\Big]}\geq \eps/4\]
\[\calB_2:\abs{\frac{1}{n}\sum_{(x,y)\in S'_{clean}}\max_{u\in\calU(x)}\frac{1}{m}\sum_{z\in u\cap S'_{perturb}}\ind [h(z)\neq y] - \Ex_{(x,y)\sim D} \Big[\max_{u\in\calU(x)}\frac{1}{m}\sum_{z\in u\cap S'_{perturb}}\ind [h(z)\neq y]\Big]}\geq \eps/4\]

Putting~\Cref{eqn:hoeffding-5,eqn:hoeffding-6} together, 
\begin{align}
&\Pr_{S'_{clean}\sim \calD^n, r\sim\calR}\Bigg[\abs{\frac{1}{n}\sum_{(x,y)\in S'_{clean}}\max_{u\in\calU(x)}\frac{1}{m}\sum_{z\in u\cap S'_{perturb}}\ind [h(z)\neq y] - \Ex_{(x,y)\sim \calD}\Big[\max_{u\in\calU(x)}\Ex_{z\sim u}\ind [h(z)\neq y]\Big]}\geq \eps/2\Bigg]\nonumber\\
\leq&\Pr_{S'_{clean}\sim \calD^n, r\sim\calR}[\calB_1=1]+\Pr_{S'_{clean}\sim \calD^n, r\sim\calR}[\calB_2=1]\nonumber\\
\leq& 54/100+\Pr_{r\sim \calR}[r]\cdot\Pr_{S'_{clean}\sim \calD^n}[B_2=1|r]\nonumber\\
\leq&54/100+6/100=3/5\label{eqn:hoeffding-7}
\end{align}

where~\Cref{eqn:hoeffding-7} holds for all $h\in \calH$. Now, $\Pr(B)\geq \Pr(A,B) = \Pr(A)\Pr(B| A)$. Consider drawing set $S$ and suppose event $A$ occurs. Let $h$ be in $\calH$ such that:

\[\Ex_{(x,y)\sim \calD} \insquare{ \max_{u\in \calU(x)}\insquare{\Ex_{z\sim u} \ind\insquare{h(z)\neq y}} }\geq \eps , \frac{1}{n}\sum_{(x,y)\in S_{clean}} \insquare{ \max_{u\in \calU(x)}\insquare{\frac{1}{m}\sum_{z\in \calU(x)\cap S_{perturb}} \ind\insquare{h(z)\neq y}} }=0\]

By~\Cref{eqn:hoeffding-7}, given $\Ex_{(x,y)\sim \calD} \insquare{ \max_{u\in \calU(x)}\insquare{\Ex_{z\sim u} \ind\insquare{h(z)\neq y}} }\geq \eps$, when $m=\Omega(1/\eps^2\cdot\log(1/\eps)),n=\Omega(1/\eps^2)$, %
\[\Pr\Bigg[\frac{1}{n}\sum_{(x,y)\in S'_{clean}}\max_{u\in\calU(x)}\frac{1}{m}\sum_{z\in u\cap S'_{perturb}}\ind [h(z)\neq y] \leq \eps/2\Bigg]\leq 3/5\].

Thus, $\Pr(B|A)\geq 2/5$ and $\Pr(B)\geq (2/5)\Pr(A)$ as desired.
\end{proof}

\subsection{Proof of~\Cref{thm:robust-loss-realizable-pre-Sauer}}
\begin{proof}
Consider drawing a set $S_{c}$ of $n$ examples from $\calD$ and then for each $(x,y)\in S_{c}$, add $m$ perturbations sampled from each $u\in \calU(x)$ to $S_{p}$ and let $S=S_{c}\cup S_{p}$.
Define $A$ as the event where there exists $h\in\calH$ with $\DR_{\calD}(h)>\eps$ but $\DR_S(h)=0$. By~\Cref{lem:robust-loss}, it suffices to prove that 
$\Pr(B)\leq (2/5)\delta$, where $B$ is the same event as defined in~\Cref{lem:robust-loss}.
Consider a third experiment. Draw a set $S''_{c}$ of $2n$ points from $\calD$ and then augment each natural example $(x,y)$ by adding $mk$ perturbations sampled from each $u\in\calU(x)$ to get a set $S''$. Now, in fact, the set $S''=\{B_1,\cdots,B_{2n}\}$ where each $B_i\in S''$ is a ball around the $i^{th}$ clean example in $S''_{c}$ that contains the $i^{th}$ clean example and all its $m\cdot k$ perturbations. Next, randomly partition $S''$ into two sets $S$ and $S'$ of $n$ balls each.

Let $B^*$ denote the event that there exists $h\in\calH$ with $\DR_S(h)=0$ but $\DR_{S'}(h)\geq \eps/2$. $\Pr(B^*)=\Pr(B)$ since drawing $2n$ points from $\calD$ then augmenting them by adding $m\cdot k$ perturbations per each natural example, and randomly partitioning them into two sets of size $n$, results in the same distribution on $(S,S')$ as does drawing $S$ and $S'$ directly. %
The advantage of this new experiment is that we can now argue that $\Pr(B^*)$ is low, with probability now taken over just the random partition of $S''$ into $S$ and $S'$. The key point is that since $S''$ is fixed, there are at most $|\calH[S'']|\leq \calH[2n\cdot m \cdot k]$ events to worry about. Specifically, it suffices to prove that for any fixed $h\in \calH[S'']$, the probability over the partition of $S''$ that $h$ makes zero distributional adversarial loss on $S$ but $\DR_{S'}(h)\geq \eps/2$
is at most %
$2\delta/(5\calH[2n\cdot m\cdot k])$.We can then apply the union bound.

Consider the following specific method for partitioning $S''$ into $S$ and $S'$. Randomly put the balls in $S''$ into pairs: $(a_1,b_1),(a_2,b_2),\cdots,(a_n,b_n)$. For each index $i$, flip a fair coin. If heads put $a_i$ in $S$ and $b_i$ into $S'$, else if tails put $a_i$ into $S'$ and $b_i$ into $S$. Now, fix some partition $h\in \calH[S'']$ and consider the probability over these $n$ fair coin flips such $\DR_S(h)=0$ that %
but $\DR_{S'}(h)\geq \eps/2$.
First of all, if for any index $i$, $h$ makes a robustness mistake, i.e. a mistake on any examples inside a ball, on both $a_i$ and $b_i$ then the probability is zero (because it cannot have zero robust loss on $S$). Second, if there are fewer than $\eps n/2$ indices $i$ such that $h$ makes a robustness mistake on either $a_i$ or $b_i$ then again the probability is zero because it cannot possibly %
be the case that $\DR_{S'}(h)\geq \eps/2$. So, assume there are $r\geq \eps n/2$ indices $i$ such that $h$ makes a robustness mistake on exactly one of $a_i$ or $b_i$. In this case, the chance that all of those mistakes land in $S'$ is exactly $1/2^r$. This quantity is at most $1/2^{\eps n/2}\leq 2\delta/(5\calH[2n\cdot m \cdot k])$ as desired for $n$ given in the theorem statement.
\end{proof}

\subsection{Proof of~\Cref{thm:vc-dim-robustly-realizable} }
\begin{proof}%
We use Sauer's Lemma and%
~\Cref{thm:robust-loss-realizable-pre-Sauer} to complete the proof.

Using Sauer's lemma, we have that for $\calH[2n\cdot m\cdot k]\leq (2enmk/d)^d$. From%
~\Cref{thm:robust-loss-realizable-pre-Sauer} we have the following:
\[n\geq\frac{2}{\eps}[\log(2\calH[2n\cdot m\cdot k])+\log(\frac{1}{\delta})+\frac{7}{\eps}]\]
Combining with Sauer's lemma implies that:
\[n\geq\frac{2}{\eps}[\log(2(\frac{2enmk}{d})^d)+\log(\frac{1}{\delta})+\frac{7}{\eps}]\]
Therefore,
\[n\geq \frac{2}{\eps}\Bigl[d\log(n)+d\log(2emk/d)+d\log(2)+\log(1/\delta)+\frac{7}{\eps}\Bigr]\]
Therefore, it is sufficient to have: %
\begin{align}
&n\geq \frac{2}{\eps}\Bigl[d\log(n)+d\log(2emk/d)+d+\log(1/\delta)+\frac{7}{\eps}\Bigr]\\
&n\geq \frac{2d}{\eps}\log(n)+\frac{2d}{\eps}\log(2emk/d)+\frac{2d}{\eps}+\frac{2}{\eps}\log(1/\delta)+\frac{14}{\eps^2}\label{eq:sample-complexity1}
\end{align}
In order to have $x\geq a\log(x)+b$ it is sufficient to have $x\geq 4a\log(2a)+2b$. Therefore, in order to have~\Cref{eq:sample-complexity1}, it is sufficient to have:
\[n\geq \frac{8d}{\eps}\log(\frac{8d}{\eps})+\frac{4d}{\eps}\log(2emk/d)+\frac{4d}{\eps}+\frac{4}{\eps}\log(1/\delta)+\frac{28}{\eps^2}\]
And it is sufficient to have:
\[n\geq \frac{8d}{\eps}\log(\frac{16demk}{d\eps})+\frac{4d}{\eps}+\frac{4}{\eps}\log(1/\delta)+\frac{28}{\eps^2}\]
Therefore:
\[n=\calO\Big(\frac{1}{\eps}\cdot d \log(\frac{mk}{\eps})+\frac{1}{\eps}\log(\frac{1}{\delta})+\frac{1}{\eps^2}\Big)\]
\end{proof}

\subsection{Proof of~\Cref{thm:vc-dim-robustly-agnostic}}
Putting together~\Cref{lem:robust-loss-agnostic,thm:robust-loss-agnostic-pre-Sauer} proves~\Cref{thm:vc-dim-robustly-agnostic} holds.
Similar to~\Cref{lem:robust-loss}, we prove the following lemma in the agnostic case.
\begin{lemma}
\label{lem:robust-loss-agnostic}
Let $\calH$ be a concept class over a domain $\calX$. Let $S_{c}$ and $S'_{c}$ be
sets of $n$ elements drawn from some distribution $\calD$ over $\calX$, where %
$n\geq 13/\eps^2$. 
For each $(x,y)\in S_{c}$, %
$m=\Omega(1/\eps^2\log(1/\eps))$ perturbations sampled from each $u\in\calU(x)$ are added to a set $S_{p}$ and finally $S=S_{c}\cup S_{p}$. 
Similarly, $S'_{c}$ is augmented to get $S'$.
Let $A$ denote the event that there exists $h\in \calH$ such that $\abs{\DR_{\calD}(h)-\DR_S(h)}\geq \eps$. Let $B$ denote the event that there exists $h\in\calH$ such that $\abs{\DR_{S'}(h)-\DR_S(h)}\geq \eps/2$. Then $\Pr(B)\geq (2/5)\Pr(A)$.
\end{lemma}

\begin{proof}
We need to follow an approach similar to the proof of~\Cref{lem:robust-loss} with a minor modification in the last step. Similar to ~\Cref{eqn:hoeffding-7}, we have for all $h\in \calH$,
\begin{align}
&\Pr_{S'_{c}\sim \calD^n,\atop r\sim\calR}\Bigg[\abs{\frac{1}{n}\sum_{(x,y)\in S'_{c}}\max_{u\in\calU(x)}\frac{1}{m}\sum_{z\in u\cap S'_{p}}\ind [h(z)\neq y] - \Ex_{(x,y)\sim \calD}\Big[\max_{u\in\calU(x)}\Ex_{z\sim u}\ind [h(z)\neq y]\Big]}\leq \eps/2\Bigg]\nonumber\geq 2/5\\
\end{align}
which is same as saying:
\begin{align}
&\Pr\Big[\abs{\DR_{\calD}(h)-\DR_{S'}(h)}\leq \eps/2\Big]\geq 2/5
\label{inequality:lem-robust-loss-agnostic}
\end{align}

Now, $\Pr(B)\geq \Pr(A,B) = \Pr(A)\Pr(B| A)$. Consider drawing set $S$ and suppose event $A$ occurs, let $h$ be in $\calH$ such that $\abs{\DR_{\calD}(h)-\DR_S(h)}\geq \eps$. By triangle's inequality and~\Cref{inequality:lem-robust-loss-agnostic}, $\Pr(B| A)=\Pr[\abs{\DR_{S'}(h)-\DR_S(h)}\geq \eps/2 | A]\geq 2/5$. Therefore, we can conclude that $\Pr(B)\geq (2/5)\Pr(A)$.

\end{proof}
\begin{theorem}
\label{thm:robust-loss-agnostic-pre-Sauer}
For any class $\calH$ and distribution $\calD$, a training sample $S_{c}$ of size 
\[n\geq\frac{13}{\eps^2}[\log(5\calH[2n\cdot m\cdot k])+\log(\frac{1}{\delta})]\]
where for each $(x,y)\in S_{c}$,  $m=\Omega(1/\eps^2\log(1/\eps))$ %
perturbations are sampled from each of the distributions $u\in\calU(x)$. %
Let $S_{p}$ denote the set of all perturbations, then $S=S_{c}\cup S_{p}$. Let $k=\max_{x\in \calX}\calU(x)$. 
Then with probability at least $1-\delta$, every $h\in \calH$ will have $\abs{\DR_{\calD}(h)-\DR_{S}(h)}\leq \eps$.
\end{theorem}
\begin{proof}[Proof of~\Cref{thm:robust-loss-agnostic-pre-Sauer}]

By~\Cref{lem:robust-loss-agnostic}, it suffices to prove $\Pr(B)\leq (2/5)\delta$. Consider a third experiment.
Draw a set $S''_{c}$ of $2n$ points from $\calD$ and then augment each natural example $(x,y)$ by adding $m\cdot k$ perturbations sampled from each $u\in\calU(x)$ to get a set $S''$. Now, in fact, the set $S''=\{B_1,\cdots,B_{2n}\}$ where each $B_i\in S''$ is a ball around the $i^{th}$ clean example in $S''_{c}$ that contains the $i^{th}$ clean example and all its $m\cdot k$ perturbations. Next, randomly partition $S''$ into two sets $S$ and $S'$ of $n$ balls each.

Let $B^*$ be the event that there exists $h\in \calH[S'']$ such that %
$\abs{\DR_S(h)-\DR_{S'}(h)}>\eps/2$. Consider an experiment where we randomly put the balls in $S''$ into pairs $(a_i, b_i)$. For each index $i$, flip a fair coin. If heads put $a_i$ in $S$ and $b_i$ into $S'$, else if tails put $a_i$ into $S'$ and $b_i$ into $S$. Consider the 
the value of $\DR_{S'}-\DR_{S}$ and see how it changes as we flip coins for $i=1,\cdots,n$. Initially, the difference is zero. 
For a fixed pair $(a_i,b_i)$, suppose the difference between $\DR_{a_i}(h)-\DR_{b_i}(h)=\eta$ for some value of $\eta$ between $[-1,1]$. When the $i^{th}$ random coin is flipped, the difference $\DR_{S'}(h)-\DR_{S}(h)$ increases by $\eta$ with probability $1/2$ and decreases by $\eta$ with probability $1/2$. The probability that when taking a random walk of $n$ steps where each step has a length of at most $1$, we end up more than $\eps n/2$ steps away from the origin, is at the most the probability that among $n$ coin flips the number of heads differs from its expectation by more than $\eps n/4$. By Hoeffding bounds, this is at most $2e^{-\eps^2n/8}$. This quantity is at most %
$2\delta/5\calH[2nmk]$ as desired for $n\geq (8/\eps^2)(\log(5\calH[2nmk])+\log(1/\delta))$. By applying union bound, $\Pr(B)\leq (2/5)\delta$. By applying~\Cref{lem:robust-loss-agnostic}, since $n\geq 13/\eps^2$ and $m=\Omega(1/\eps^2 \log(1/\eps))$, $\Pr(A)\leq (5/2)\Pr(B)$ which implies that $\Pr(A)\leq \delta$.
\end{proof}

\begin{proof}[Proof of~\Cref{thm:vc-dim-robustly-agnostic}]
We use Sauer's Lemma and%
~\Cref{thm:robust-loss-agnostic-pre-Sauer} to complete the proof.

Using Sauer's lemma, we have that for $\calH[2n\cdot m\cdot k]\leq (2enmk/d)^d$. From%
~\Cref{thm:robust-loss-agnostic-pre-Sauer} we have the following:
\[n\geq\frac{13}{\eps^2}[\log(2\calH[2n\cdot m\cdot k])+\log(\frac{1}{\delta})]\]
Combining with Sauer's lemma implies that:
\[n\geq\frac{13}{\eps^2}[\log(2(\frac{2enmk}{d})^d)+\log(\frac{1}{\delta})]\]
Therefore,
\[n\geq \frac{13}{\eps^2}\Bigl[d\log(n)+d\log(2emk/d)+d\log(2)+\log(1/\delta)\Bigr]\]
Therefore, it is sufficient to have: %
\begin{align}
&n\geq \frac{13}{\eps^2}\Bigl[d\log(n)+d\log(2emk/d)+d+\log(1/\delta)\Bigr]\\
&n\geq \frac{13d}{\eps^2}\log(n)+\frac{13d}{\eps^2}\log(2emk/d)+\frac{13d}{\eps^2}+\frac{13}{\eps^2}\log(1/\delta)\label{eq:sample-complexity1}
\end{align}
In order to have $x\geq a\log(x)+b$ it is sufficient to have $x\geq 4a\log(2a)+2b$. Therefore, in order to have~\Cref{eq:sample-complexity1}, it is sufficient to have:
\[n\geq \frac{52d}{\eps^2}\log(\frac{26d}{\eps^2})+\frac{26d}{\eps^2}\log(2emk/d)+\frac{26d}{\eps^2}+\frac{26}{\eps^2}\log(1/\delta)\]
And it is sufficient to have:
\[n\geq \frac{52d}{\eps^2}\log(\frac{52demk}{d\eps^2})+\frac{26d}{\eps^2}+\frac{26}{\eps^2}\log(1/\delta)\]
Therefore:
\[n=\calO\Big(\frac{1}{\eps^2}\cdot d \log(\frac{mk}{\eps})+\frac{1}{\eps^2}\log(\frac{1}{\delta})\Big)\]
\end{proof}

\subsection{Proof of~\Cref{thm:vc-dim-robustly-realizable-model-representative} and extension to the agnostic case~\Cref{thm:vc-dim-robustly-agnostic-model-representative}}

\begin{proof}[Proof of~\Cref{thm:vc-dim-robustly-realizable-model-representative}]
In order to prove~\Cref{thm:vc-dim-robustly-realizable-model-representative}, they key idea is to prove~\Cref{lem:robust-loss-representative-1}, which states the following: Let $A$ denote the event that there exists $h\in \calH$ with zero empirical distributional adversarial error on $S$ (with respect to the representative perturbation sets $\calR(.)$) but true distributional adversarial error at least $\eps$ (with respect to the true perturbation sets $\calU(.)$). Now draw a \emph{fresh} test set $S'$ of $n$ examples from $\calD$ where for each example $(x,y)\in S'$, $m$ perturbations are drawn from each distribution $u\in \calR(x)$ and are added to $S'$. Let $B$ denote the event that there exists $h\in\calH$ with zero distributional adversarial loss on $S$ but distributional adversarial loss at least $\eps/2$ on $S'$. We prove that $\Pr(B)\geq (2/5)\Pr(A)$.

\begin{lemma}
\label{lem:robust-loss-representative-1}
Let $\calH$ be a concept class over a domain $\calX$. Let $S_{c}$ and $S'_{c}$ be
sets of $n$ elements drawn from some distribution $\calD$ over $\calX$, where $n=\Omega(1/\eps^2)$. %
For each $(x,y)\in S_{c}$, %
$m=\Omega(1/\eps^2\log(1/\eps))$ perturbations sampled from each $u\in\calR(x)$ are added to a set $S_{p}$ and finally $S=S_{c}\cup S_{p}$. 
Similarly, $S'_{c}$ is augmented to get $S'$.
Let $A$ denote the event that there exists $h\in \calH$ with zero empirical distributional adversarial error on $S$ but true distributional adversarial error $\geq \eps+\eps'$:
\begin{align*}
&\Ex_{(x,y)\sim \calD} \insquare{ \max_{u\in \calU(x)}\insquare{\Ex_{z\sim u} \ind\insquare{h(z)\neq y}} }\geq \eps+\eps' ,
\quad\frac{1}{n}\sum_{(x,y)\in S_{c}} \insquare{ \max_{u\in \calR(x)}\insquare{\frac{1}{m}\sum_{z\in \calR(x)\cap S_{p}} \ind\insquare{h(z)\neq y}} }=0
\end{align*}
Let $B$ denote the event that there exists $h\in\calH$ with zero distributional adversarial loss on $S$ but distributional adversarial loss $\geq \eps/2$ on $S'$:
\begin{align*}
&\frac{1}{n}\sum_{(x,y)\in S'_{c}} \insquare{ \max_{u\in \calR(x)}\insquare{\frac{1}{m} \sum_{z\in \calR(x)\cap S'_{p}}\ind\insquare{h(z)\neq y}} }\geq \eps/2,\\
&\frac{1}{n}\sum_{(x,y)\in S_{c}} \insquare{ \max_{u\in \calR(x)}\insquare{\frac{1}{m}\sum_{z\in \calR(x)\cap S_{p}} \ind\insquare{h(z)\neq y}} }=0
\end{align*}
Then $\Pr(B)\geq (2/5)\Pr(A)$.
\end{lemma}

\begin{proof}[Proof of~\Cref{lem:robust-loss-representative-1}]
Suppose event $A$ happens, and let $h$ be in $\calH$ such that $\Ex_{(x,y)\sim \calD} \insquare{ \max_{u\in \calU(x)}\insquare{\Ex_{z\sim u} \ind\insquare{h(z)\neq y}} }\geq \eps+\eps'$. Let $f$ be a function that maps an input $(x,y)$ to a distribution $u\in \calU(x)$ with maximum error given classifier $h$:
\[f:(x,y)\rightarrow u\in \calU(x):u=\argmax_{\calU(x)}\Ex_{z\sim u}\ind[h(z)\neq y] \]
, and 
\[f':(x,y)\rightarrow r\in \calR(x): r=\argmin_{\calR(x)} TV(r,f(x,y))\]

Given event $A$, we know:
\begin{align}
\eps+\eps'\leq \Ex_{(x,y)\sim \calD}  \insquare{\Ex_{z\sim f(x,y)} \ind\insquare{h(z)\neq y}}%
\leq \Ex_{(x,y)\sim \calD}  \insquare{\Ex_{z\sim f'(x,y)} \ind\insquare{h(z)\neq y}}+\eps'\label{eqn:eq3-representative-1}
\end{align}

Therefore:
\begin{align}
\Ex_{(x,y)\sim \calD}  \insquare{\Ex_{z\sim f'(x,y)} \ind\insquare{h(z)\neq y}}\geq \eps%
\end{align}
which implies that:
\begin{align}
\Ex_{(x,y)\sim \calD} \insquare{ \max_{u\in \calR(x)}\insquare{\Ex_{z\sim u} \ind\insquare{h(z)\neq y}} }\geq \eps\label{eqn:eq3-representative-1}
\end{align}
Combining~\Cref{eqn:eq3-representative-1} with~\Cref{lem:robust-loss} implies that event $B$ happens with probability at least $(2/5) \Pr(A)$. %
\end{proof}

The rest of the proof of~\Cref{thm:vc-dim-robustly-realizable-model-representative} is similar to the proof of~\Cref{thm:realizable-sample-complexity}. First, we use the application of~\Cref{lem:robust-loss-representative-1} to argue about the distributional adversarial loss on the test data with perturbations generated from representative distributions $\calR(.)$ instead of population distributional adversarial loss with respect to $\calU(.)$. Furthermore, similar to the proof of~\Cref{thm:robust-loss-realizable-pre-Sauer}, we can show that for large enough test and training data coming from the same distribution, it cannot be the case that there is a large gap between training and test distributional adversarial loss. In the end, the application of Sauer's lemma similar to the proof of~\Cref{thm:realizable-sample-complexity} completes the proof.
\end{proof}

\begin{theorem}[agnostic case]
For any class $\calH$ and distribution $\calD$, a training sample $S_{c}$ of size $n=\calO\Big(\frac{1}{\eps^2}[VCdim(\calH)\log(\frac{mk}{\eps})+\log(\frac{1}{\delta})]\Big)$, where for each $(x,y)\in S_{c}$,  
$m=\Omega(1/\eps^2\log(1/\eps))$ perturbations are sampled from each of the distributions $u\in\calR(x)$. Let $S_{p}$ denote the set of all perturbations, then $S=S_{c}\cup S_{p}$. Given sample set $S$, with probability $\geq 1-\delta$, for every $h\in \calH$, $\abs{\DR_{\calD}(h)-\DR_{S}(h)}\leq \eps'+\eps$.
\label{thm:vc-dim-robustly-agnostic-model-representative}
\end{theorem}

\begin{proof}[Proof of~\Cref{thm:vc-dim-robustly-agnostic-model-representative}]
In the agnostic case, first similar to~\Cref{lem:robust-loss-representative-1}, we can show the following lemma holds.
\begin{lemma}
\label{lem:robust-loss-agnostic-representative-1}
Let $\calH$ be a concept class over a domain $\calX$. Let $S_{c}$ and $S'_{c}$ be
sets of $n$ elements drawn from some distribution $\calD$ over $\calX$, where %
$n\geq 13/\eps^2$. 
For each $(x,y)\in S_{c}$, %
$m=\Omega(1/\eps^2\log(1/\eps))$ perturbations sampled from each $u\in\calR(x)$ are added to a set $S_{p}$ and finally $S=S_{c}\cup S_{p}$. 
Similarly, $S'_{c}$ is augmented to get $S'$.
Let $A$ denote the event that there exists $h\in \calH$ such that $\abs{\DR_{\calD}(h)-\DR_S(h)}\geq \eps+\eps'$. Let $B$ denote the event that there exists $h\in\calH$ such that $\abs{\DR_{S'}(h)-\DR_S(h)}\geq \eps/2$. Then $\Pr(B)\geq (2/5)\Pr(A)$.
\end{lemma}

The rest of the proof of~\Cref{thm:vc-dim-robustly-agnostic-model-representative} is similar to the proof of~\Cref{thm:agnostic-sample-complexity}. First, we use the application of~\Cref{lem:robust-loss-agnostic-representative-1} to argue about the distributional adversarial loss on the test data with perturbations generated from representative distributions $\calR(.)$ instead of population distributional adversarial loss with respect to $\calU(.)$. Furthermore, similar to the proof of~\Cref{thm:robust-loss-agnostic-pre-Sauer}, we can show that for large enough test and training data coming from the same distribution, it cannot be the case that there is a large gap between training and test distributional adversarial loss. In the end, the application of Sauer's lemma similar to the proof of~\Cref{thm:agnostic-sample-complexity} completes the proof.
\end{proof}

\subsection{Proof of~\Cref{thm:vc-dim-robustly-realizable-extension-2} and extension to the agnostic case}

\begin{proof}[Proof of~\Cref{thm:vc-dim-robustly-realizable-extension-2}]
In order to prove~\Cref{thm:vc-dim-robustly-realizable-extension-2} holds, first we prove the following lemma:
\begin{lemma}
\label{lem:robust-loss-representative-2}
Let $\calH$ be a concept class over a domain $\calX$. Let $S_{c}$ and $S'_{c}$ be
sets of $n$ elements drawn from some distribution $\calD$ over $\calX$, where $n=\Omega(1/\eps^2)$.%
For each $(x,y)\in S_{c}$, %
$m=\Omega(1/\eps^2\log(1/\eps))$ perturbations sampled from each $u\in\calR(x)$ are added to a set $S_{p}$ and finally $S=S_{c}\cup S_{p}$. Similarly, $S'_{c}$ is augmented to get $S'$.
Let $A$ denote the event that there exists $h\in \calH$ with zero empirical distributional adversarial error on $S$ but true distributional adversarial error $\geq k\eps$:
\begin{align*}
&\Ex_{(x,y)\sim \calD} \insquare{ \max_{u\in \calU(x)}\insquare{\Ex_{z\sim u} \ind\insquare{h(z)\neq y}} }\geq k\eps ,\\
&\frac{1}{n}\sum_{(x,y)\in S_{c}} \insquare{ \max_{u\in \calR(x)}\insquare{\frac{1}{m}\sum_{z\in \calR(x)\cap S_{p}} \ind\insquare{h(z)\neq y}} }=0
\end{align*}
Let $B$ denote the event that there exists $h\in\calH$ with zero distributional adversarial loss on $S$ but distributional adversarial loss $\geq \eps/2$ on $S'$:
\begin{align*}
&\frac{1}{n}\sum_{(x,y)\in S'_{c}} \insquare{ \max_{u\in \calR(x)}\insquare{\frac{1}{m} \sum_{z\in \calR(x)\cap S'_{p}}\ind\insquare{h(z)\neq y}} }\geq \eps/2 ,\\
&\frac{1}{n}\sum_{(x,y)\in S_{c}} \insquare{ \max_{u\in \calR(x)}\insquare{\frac{1}{m}\sum_{z\in \calR(x)\cap S_{p}} \ind\insquare{h(z)\neq y}} }=0
\end{align*}
Then $\Pr(B)\geq (2/5)\Pr(A)$.
\end{lemma}

\begin{proof}[Proof of~\Cref{lem:robust-loss-representative-2}]
Suppose event $A$ happens, and let $h$ be in $\calH$ such that $\Ex_{(x,y)\sim \calD} \insquare{ \max_{u\in \calU(x)}\insquare{\Ex_{z\sim u} \ind\insquare{h(z)\neq y}} }\geq k\eps$. Let $f$ be a function that maps an input $(x,y)$ to a distribution $u\in \calU(x)$ with maximum error given classifier $h$:
\[f:(x,y)\rightarrow u\in \calU(x):u=\argmax_{\calU(x)}\Ex_{z\sim u}\ind[h(z)\neq y], \]
and 
{\small\[f':(x,y)\rightarrow r\in \calR(x): r\text{ has the maximum coverage of the error region of } f(x,y)\].}

Given event A, we know:
\begin{align}
&\eps\leq \frac{1}{k}\Ex_{(x,y)\sim \calD}  \insquare{\Ex_{z\sim f(x,y)} \ind\insquare{h(z)\neq y}}
\leq \Ex_{(x,y)\sim \calD}  \insquare{\Ex_{z\sim f'(x,y)} \ind\insquare{h(z)\neq y}}
\end{align}

where the last inequality holds by the pigeon-hole principle.
Therefore $\Ex_{(x,y)\sim \calD}  \insquare{\Ex_{z\sim f'(x,y)} \ind\insquare{h(z)\neq y}}\geq \eps$, which implies that:
\[\Ex_{(x,y)\sim \calD} \insquare{ \max_{u\in \calR(x)}\insquare{\Ex_{z\sim u} \ind\insquare{h(z)\neq y}} }\geq \eps\]
Combining it with~\Cref{lem:robust-loss} implies that event $B$ happens with probability at least $(2/5)\Pr(A)$. 
\end{proof}

The rest of the proof of~\Cref{thm:vc-dim-robustly-realizable-extension-2} is similar to~\Cref{thm:realizable-sample-complexity}. First, we use the application of~\Cref{lem:robust-loss-representative-2} to argue about the distributional adversarial loss on the test data with perturbations generated from representative distributions $\calR(.)$ instead of population distributional adversarial loss with respect to $\calU(.)$. Furthermore, similar to the proof of~\Cref{thm:robust-loss-realizable-pre-Sauer}, we can show that for large enough test and training data coming from the same distribution, it cannot be the case that there is a large gap between training and test distributional adversarial loss. In the end, the application of Sauer's lemma similar to the proof of~\Cref{thm:realizable-sample-complexity} completes the proof.
\end{proof}

\begin{theorem}%
[agnostic case]
For any class $\calH$ and distribution $\calD$, a training sample $S_{c}$ of size $n=\calO\Big(\frac{1}{\eps^2}[VCdim(\calH)\log(\frac{mk}{\eps})+\log(\frac{1}{\delta})]\Big)$, where for each $(x,y)\in S_{c}$,  
$m=\Omega(1/\eps^2\log(1/\eps))$ perturbations are sampled from each of the distributions $u\in\calR(x)$. Let $S_{p}$ denote the set of all perturbations, then $S=S_{c}\cup S_{p}$. Given sample set $S$, %
with probability $\geq 1-\delta$, for every $h\in \calH$, $\abs{\DR_{\calD}(h)-\DR_{S}(h)}\leq k\eps$.
\label{thm:vc-dim-robustly-agnostic-extension-2}
\end{theorem}
\begin{proof}[Proof of~\Cref{thm:vc-dim-robustly-agnostic-extension-2}]

In the agnostic case, first similar to~\Cref{lem:robust-loss-representative-2} we can show the following lemma holds:
\begin{lemma}
\label{lem:robust-loss-agnostic-representative-2}
Let $\calH$ be a concept class over a domain $\calX$. Let $S_{c}$ and $S'_{c}$ be
sets of $n$ elements drawn from some distribution $\calD$ over $\calX$, where %
$n\geq 13/\eps^2$. 
For each $(x,y)\in S_{c}$, %
$m=\Omega(1/\eps^2\log(1/\eps))$ perturbations sampled from each $u\in\calR(x)$ are added to a set $S_{p}$ and finally $S=S_{c}\cup S_{p}$. 
Similarly, $S'_{c}$ is augmented to get $S'$.
Let $A$ denote the event that there exists $h\in \calH$ such that $\abs{\DR_{\calD}(h)-\DR_S(h)}\geq k\eps$. Let $B$ denote the event that there exists $h\in\calH$ such that $\abs{\DR_{S'}(h)-\DR_S(h)}\geq \eps/2$. Then $\Pr(B)\geq (2/5)\Pr(A)$.
\end{lemma}

The rest of the proof of~\Cref{thm:vc-dim-robustly-agnostic-extension-2} is similar to the proof~\Cref{thm:agnostic-sample-complexity}. First, we use the application of~\Cref{lem:robust-loss-agnostic-representative-2} to argue about the distributional adversarial loss on the test data with perturbations generated from representative distributions $\calR(.)$ instead of population distributional adversarial loss with respect to $\calU(.)$. Furthermore, similar to the proof of~\Cref{thm:robust-loss-agnostic-pre-Sauer}, we can show that for large enough test and training data coming from the same distribution, it cannot be the case that there is a large gap between training and test distributional adversarial loss. In the end, the application of Sauer's lemma similar to the proof of~\Cref{thm:agnostic-sample-complexity} completes the proof.
\end{proof}

\subsection{Proof of \autoref{thm:derandomization_classifier} }
\begin{proof}
    Fix a sample $(x,y)$ such that $\epsilon(x,y)\leq 1/2-\eta$. We will show that 
    \[
    \Pr_{R_1,R_2,\ldots,R_t}[\exists x'\in \calA(x)\colon h^{(R_1,R_2,\ldots,R_t)}(x')\neq y ]\leq \delta.
    \]
    For this fix we fix an $x'\in \calA(x)$ and analyze 
    \[
    \Pr_{R_1,R_2,\ldots,R_t}[h^{(R_1,R_2,\ldots,R_t)}(x')\neq y].
    \]
    Notice that $\Pr_{R\sim \calR}[h(x',R)\neq y]\leq 1/2-\eta$ and therefore by the Chernoff bound we have 
    \begin{align*}
             \Pr_{R_1,R_2,\ldots,R_t}[h^{(R_1,R_2,\ldots,R_t)}(x')\neq y]\leq 
      \Pr_{R_1,R_2,\ldots,R_t}[\sum_{i=1}^t \ind[h(x,R_i)\neq y]\geq t/2]
      \leq \exp{(-2\eta^2 t)}.
    \end{align*}

    Finally, by a union bound over the set $\calA(x)$ we have 
    \begin{align*}
            \Pr_{R_1,R_2,\ldots,R_t}[\exists x'\in \calA(x)\colon h^{(R_1,R_2,\ldots,R_t)}(x')\neq y ]\leq |\calA(x)|\cdot \exp{(-2\eta^2 t)} \leq \delta^2.
    \end{align*}
    Since, this is true for any fixed $(x,y)$ such that $\epsilon(x,y)\leq 1/2-\eta$, we have
    \begin{align*}
      \E\limits_{\substack{(x,y)\sim D\\ R_1,R_2,\ldots,R_t}} [\exists x'\in \calA(x)\colon h^{(R_1,R_2,\ldots,R_t)}(x')\neq y | \epsilon(x,y)<1/2-\eta] \leq \delta^2,\\
        \intertext{ and by Markov's inequality this yields}
        \Pr_{R_1,R_2\ldots,R_t}\insquare{\E_{(x,y)\sim \calD}[\exists x'\in \calA(x)\colon h^{(R_1,R_2,\ldots,R_t)}(x')\neq y | \epsilon(x,y)<1/2-\eta ]\geq \delta} \leq \delta.
    \end{align*}
    
    Our desired claim follows by noting that when $(x,y)$ is sampled from $\calD$, then with probability at most $\epsilon(\eta)$ we have $\epsilon(x,y)>1/2-\eta$:
    \begin{align*}
            &\Pr_{R_1,R_2\ldots,R_t}\insquare{\E_{(x,y)\sim \calD}[\exists x'\in \calA(x)\colon h^{(R_1,R_2,\ldots,R_t)}(x')\neq y ]\geq \epsilon(\eta)+\delta} \leq\\
             &\Pr_{R_1,R_2\ldots,R_t}\insquare{\epsilon(\eta)+\E_{(x,y)\sim \calD}[\exists x'\in \calA(x)\colon h^{(R_1,R_2,\ldots,R_t)}(x')\neq y | \epsilon(x,y)<1/2-\eta ]\geq \epsilon(\eta) + \delta} \leq \\
            &\Pr_{R_1,R_2\ldots,R_t}\insquare{\E_{(x,y)\sim \calD}[\exists x'\in \calA(x)\colon h^{(R_1,R_2,\ldots,R_t)}(x')\neq y | \epsilon(x,y)<1/2-\eta ]\geq \delta} \leq \delta.
    \end{align*}
\end{proof}

\subsection{Proof of \autoref{thm:derandomization_certifiable}}
\begin{proof}
    Fix a sample $(x,y)$ such that $\gamma(\rho,x,y)\leq 1/2-\eta$. We will show that 
\begin{align*}
    \Pr_{R_1,\ldots,R_t}\insquare{\exists x'\in \calA(x): \frac{\rho^{(R_1,R_2,\ldots,R_t)}(x')}{\robreg(h,x')}\notin [1-\beta,1+\alpha]} \leq \delta.
\end{align*}
For this consider a fixed $x'\in \calA(x)$. Notice that 
\begin{align*}
    \Pr_{R\sim\calR}\insquare{\frac{\rho(x',R)}{\robreg(h,x')}\notin [1-\beta,1+\alpha]} \leq 1/2-\eta.
    \end{align*}
Therefore, by the Chernoff bound we have
\begin{align*}
    \Pr_{R_1,\ldots,R_t}\insquare{ \frac{\rho^{(R_1,R_2,\ldots,R_t)}(x')}{\robreg(h,x')}\notin [1-\beta,1+\alpha]} &\leq\\ 
    \Pr_{R_1,\ldots,R_t}\insquare{\sum_{i=1}^t \ind[\frac{\rho(x',R_i)}{\robreg(h,x')}\leq 1-\beta] \geq t/2}  
     &+  
      \Pr_{R_1,\ldots,R_t}\insquare{\sum_{i=1}^t \ind[\frac{\rho(x',R_i)}{\robreg(h,x')}\geq 1+\alpha] \geq t/2} \leq 2\cdot \exp{(-2\eta^2 t)}.
\end{align*}
Finally, by a union bound over $\calA(x)$ we have
\begin{align*}
    \Pr_{R_1,\ldots,R_t}\insquare{\exists x'\in \calA(x): \frac{\rho^{(R_1,R_2,\ldots,R_t)}(x')}{\robreg(h,x')}\notin [1-\beta,1+\alpha]} \leq 
    2|\calA(x)|\cdot \exp{(-2\eta^2 t)} \leq \delta^2.
\end{align*}
Since, this is true for any fixed $(x,y)$ such that $\epsilon(x,y)\leq 1/2-\eta$, we have
    \begin{align*}
      \E\limits_{\substack{(x,y)\sim D\\ R_1,R_2,\ldots,R_t}} [\exists x'\in \calA(x)\colon \frac{\rho^{(R_1,R_2,\ldots,R_t)}(x')}{\robreg(h,x')}\notin [1-\beta,1+\alpha] | \epsilon(x,y)<1/2-\eta] \leq \delta^2,\\
        \intertext{ and by Markov's inequality this yields}
        \Pr_{R_1,R_2\ldots,R_t}\insquare{\E_{(x,y)\sim \calD}[\exists x'\in \calA(x)\colon \frac{\rho^{(R_1,R_2,\ldots,R_t)}(x')}{\robreg(h,x')}\notin [1-\beta,1+\alpha] | \epsilon(x,y)<1/2-\eta ]\geq \delta} \leq \delta.
    \end{align*}
    
    Our desired claim follows by noting that when $(x,y)$ is sampled from $\calD$, then with probability at most $\epsilon(\eta)$ we have $\epsilon(x,y)>1/2-\eta$:
    \begin{align*}
            &\Pr_{R_1,R_2\ldots,R_t}\insquare{\E_{(x,y)\sim \calD}[\exists x'\in \calA(x)\colon \frac{\rho^{(R_1,R_2,\ldots,R_t)}(x')}{\robreg(h,x')}\notin [1-\beta,1+\alpha] ]\geq \epsilon(\eta)+\delta} \leq\\
             &\Pr_{R_1,R_2\ldots,R_t}\insquare{\epsilon(\eta)+\E_{(x,y)\sim \calD}[\exists x'\in \calA(x)\colon \frac{\rho^{(R_1,R_2,\ldots,R_t)}(x')}{\robreg(h,x')}\notin [1-\beta,1+\alpha] | \epsilon(x,y)<1/2-\eta ]\geq \epsilon(\eta) + \delta} \leq \\
            &\Pr_{R_1,R_2\ldots,R_t}\insquare{\E_{(x,y)\sim \calD}[\exists x'\in \calA(x)\colon \frac{\rho^{(R_1,R_2,\ldots,R_t)}(x')}{\robreg(h,x')}\notin [1-\beta,1+\alpha] | \epsilon(x,y)<1/2-\eta ]\geq \delta} \leq \delta. 
    \end{align*}
\end{proof}

\end{document}